\documentclass{article}

\usepackage{iclr2023_conference,times}
\iclrfinalcopy

\usepackage[utf8]{inputenc} 
\usepackage[T1]{fontenc}    
\usepackage{hyperref}       
\usepackage{url}            
\usepackage{booktabs}       
\usepackage{amsfonts}       
\usepackage{nicefrac}       
\usepackage{microtype}      
\usepackage{xcolor}         
\usepackage{wrapfig}
\usepackage{bm}

\usepackage{hyperref}
\usepackage{url}
\usepackage{graphicx}
\usepackage{mathtools}
\usepackage{xspace}
\usepackage{colortbl, booktabs} 
\usepackage{subcaption}
\usepackage{tabularx}
\usepackage{array}
\usepackage{makecell}
\usepackage{enumitem}
\usepackage{overpic}

\usepackage[auth-lg,affil-sl]{authblk}

\author[1]{Nicholas Carlini\thanks{Authors ordered alphabetically.}}
\author[1,2]{Daphne Ippolito}
\author[1]{Matthew Jagielski}
\author[1,3]{\\Katherine Lee}
\author[1]{Florian Tram{\`e}r}
\author[1]{Chiyuan Zhang}

\affil[1]{Google Research}
\affil[2]{University of Pennsylvania}
\affil[3]{Cornell University}

\date{}

\usepackage{natbib}

\usepackage[utf8]{inputenc} 
\usepackage[T1]{fontenc}    
\usepackage{hyperref}       
\usepackage{url}            
\usepackage{booktabs}       
\usepackage{amsfonts}       
\usepackage{nicefrac}       
\usepackage{microtype}      
\usepackage{xcolor}         
\usepackage{amsmath,amsthm}

\theoremstyle{plain}
\newtheorem{theorem}{Theorem}[section]

\theoremstyle{definition}
\newtheorem{definition}[theorem]{Definition}

\theoremstyle{remark}

\title{Quantifying Memorization Across\\ Neural Language Models}

\definecolor{blue1}{rgb}{0,0,0.1}

\newcommand{\minisection}[1]{\noindent{\bf #1}\hspace{0.4em}}

\begin{document}
\maketitle

\begin{abstract}
Large language models (LMs) have been shown to memorize parts of their training data, and when prompted appropriately, they will emit the memorized training data verbatim. 
This is undesirable because memorization violates privacy (exposing user data), degrades utility (repeated easy-to-memorize text is often low quality), and hurts fairness (some texts are memorized over others). 

We describe three log-linear relationships that quantify the degree to which LMs emit memorized training data.
Memorization significantly grows as we increase (1) the capacity of a model, (2) the number of times an example has been duplicated, and (3) the number of tokens of context used to prompt the model. Surprisingly, we find the situation becomes more complicated when generalizing these results across model families.
On the whole, we find that memorization in LMs is more prevalent than previously believed and will likely get worse as models continues to scale, at least without active mitigations.
\end{abstract}

\section{Introduction}

The performance of neural language models has continuously improved as these models have grown from millions to trillions of parameters \citep{fedus2021switch}, with their training sets similarly growing from millions to trillions of tokens.
In anticipation of future, even larger models trained on minimally curated datasets,
it is important to 
quantify factors that lead to increased memorization of a model's training set.
%
Indeed, recent work has shown that \emph{training data extraction attacks} are a practical threat for current language models~\citep{carlini2020extracting};
an adversary interacting with a pretrained model can extract individual
sequences that were used to train the model.

While current attacks are effective, they only represent a lower bound on how much memorization occurs in existing models.
For example, by querying
the GPT-2 language model, \citet{carlini2020extracting} (manually) identified just $600$ memorized training examples out of a $40$GB training dataset.
This attack establishes a (loose) lower bound that at least
0.00000015\%
of the dataset is memorized. 
%
In contrast, we are able to show that the 6 billion parameter 
GPT-J model \citep{gpt-neo,gpt-j} \textbf{memorizes at least $\mathbf{1}\boldsymbol{\%}$ of its training dataset}: The Pile \citep{gao2020pile}.

In addition to prior work's loose estimates of models' memorization capabilities, there is a limited understanding of how memorization varies across different neural language models and datasets of different scales.
Prior studies of memorization in language models either focus on models or datasets of a fixed size~\citep{carlini2019secret, zhang2021counterfactual, thakkar2020understanding} or identify a narrow memorization-versus-scale relationship~\citep{carlini2020extracting, 2021dedup}.
While \citet{mccoy2021raven} broadly study the extent to which language models memorize, their focus is on how to avoid the problem and ensure novelty of model outputs, rather than on studying model risk through identifying the maximal amount of data memorization.

This paper addresses both of the above open questions by 
comprehensively quantifying memorization across three families of neural language models and their associated datasets.
We leverage access to each model's original training set to provide order-of-magnitude more precise bounds on the amount of extractable data that an adversary could recover than in prior works.

We first construct a set of prompts from the model's training set. 
By feeding prefixes of these prompts into the trained model,
we check whether the model has the ability to complete the rest of the example verbatim.
%
%
This allows us to measure memorization across models, datasets, and prompts of varying sizes. We identify three properties that significantly impact memorization:

\begin{enumerate}[noitemsep, topsep=-2pt, wide, labelwidth=5pt, labelindent=5pt]
    \item \textbf{Model scale:} 
    Within a model family, larger models memorize $2$-$5\times$ more than smaller models. 

    \item \textbf{Data duplication:} Examples repeated more often are more likely to be extractable.
    
    \item \textbf{Context:} It is orders of magnitude easier to extract sequences when given a longer context.
\end{enumerate}

Our analysis suggests that future research on neural language modeling will need to take steps
to prevent future (larger) models from memorizing their training datasets.

\section{Related Work}

There is extensive prior work that qualitatively studies memorization in neural language models. Prior work has demonstrated \emph{extraction attacks} that recover memorized data including URLs, phone numbers, and other personal information~\citep{carlini2020extracting, ziegler2021copilot}---or synthetically injected ``canaries''~\citep{carlini2019secret, henderson2017ethical, thakkar2020understanding, thomas2020investigating}.
However most of these works are qualitative and aim to demonstrate the existence of extractable data, rather than precisely quantifying how much models memorize.
For example, the unprompted memorization evaluation of \citet{carlini2020extracting} found just 600 examples of memorization in GPT-2.
Our paper aims to establish tighter bounds on the fraction of a dataset that is memorized.

Our analysis is relevant to the broad literature on privacy attacks on machine learning.
For example, membership inference attacks~\citep{shokri2017membership, yeom2018privacy} let an adversary detect the presence of a given example in a model's training set; other forms of data leakage let an adversary learn dataset properties \citep{ganju2018property,fredrikson2015model}.
We focus on extraction attacks due to their relevance for language modeling---extraction implies significant leakage from a model, and grows with data duplication \citep{2021dedup}, a common feature of large-scale text datasets.


Various definitions of memorization in deep neural networks have been studied in prior work~\citep{carlini2019secret,carlini2020extracting,feldman2020neural,zhang2021counterfactual}. 
A detailed comparison with those existing formulations is presented in Section~\ref{sec:mem-def}.
One leading general memorization definition is differential privacy~\citep{dwork2006calibrating}, which formalizes the idea that removing any one example from the training set should not change the trained model.
However, while differential privacy protects a single user's private information, it is ineffective for preventing memorization of highly duplicated data, and does not capture the complexity of social, linguistic data~\citep{brown2022does}.
Also, differentially private learning algorithms~\citep{abadi2016deep} generally suffer from expensive computation, slow convergence, and poor model utility, despite recent advances~\citep{anil2021large}.

In concurrent work, \citet{colinpaper} 
study how often models emit memorized data
as a function of data duplication. Their analysis
focuses on evaluating why training data extraction attacks
succeed. In contrast, we
explicitly prompt models with training data prefixes in order to measure memorization in the worst case, something that a practical attack cannot necessarily do.

\minisection{Prior scaling hypotheses.}
Our motivation to study scaling phenomena stems from anecdotal evidence in prior work that
memorization ability relates to various aspects of scale.
In particular, our analysis on model scale is informed by preliminary experiments in \citep{zhang2016understanding,carlini2020extracting},
our data duplication experiments follow in the line of \citet{2021dedup},
and our context length experiments build on hypotheses by \citet{carlini2020extracting, ziegler2021copilot}.
%
%
%
%
%
%
%

\section{Methodology}
\label{sec:method}

\subsection{Definition of Memorization}
\label{sec:mem-def}
To begin, we first select a precise definition for memorization:

\begin{definition}
A string $s$ is \emph{extractable with $k$ tokens of context} from a model $f$ if there exists a (length-$k$) string $p$, such that the concatenation $[p \mid \mid s]$ is contained in the training data for $f$, and $f$ produces $s$ when prompted with $p$ using greedy decoding.
\label{def:extractable}
\end{definition}

For example, if a model's training dataset contains the sequence \textit{``My phone number is 555-6789''},
and given the length $k=4$ prefix \textit{``My phone number is''}, the most likely output is \textit{``555-6789''}, then this sequence is extractable (with 4 words of context).
We focus on greedy sampling in this paper, and verify in Section~\ref{sec:modelsize} that our choice of decoding strategy does not significantly impact our results.

While prior work proposed other definitions, we prefer ours in this paper as it is more actionable.
Some memorization definitions, including lower-bounds on differential privacy~\citep{dwork2006calibrating, jagielski2020auditing, nasr2021adversary} or counterfactual memorization~\citep{feldman2020neural, zhang2021counterfactual}, require training hundreds or thousands of models, 
which is impractical for large language models. 
Alternatively, computing \emph{exposure} \citep{carlini2019secret} requires thousands of generations per sequence, and is only designed for carefully crafted training examples.
Finally, $k$-eidetic memorization \citep{carlini2020extracting}, 
is a useful definition for \emph{unprompted} memorization, but less useful for tightly bounding memorization by prompting with training data (as we will do).
%
%
%
%
%
%
Future work might explore how our three scaling observations apply to other definitions of memorization. 

\subsection{Selection of Evaluation Data}

Having chosen a definition, we next describe our evaluation procedure.
%
Ideally, we would consider every sequence $x=[p \mid \mid s]$ in the model's training dataset (where $x$ has been split into a length-$k$ prefix $p$ and a suffix $s$). For each sequence, we would report if the model exactly reproduces $s$ when prompted with $p$, following Definition \ref{def:extractable}.
Unfortunately, performing this test on every sequence in the training data would be prohibitively expensive.
For example, the largest 6 billion parameter GPT-Neo model has a throughput of roughly one 100-token generation per second on a V100 GPU. Extrapolating to the 800GB training dataset, this would require over $30$ GPU-years of compute.

Instead, we query on a smaller subset of the training data, that still produces statistically confident estimates.
In this paper we randomly choose subsets of roughly $50{,}000$ sequences,
allowing us to efficiently run inference in just a few hours.
The primary criteria when choosing a subset of the training data is to obtain a representative sample that allows
us to draw meaningful conclusions from the data.
We consider two approaches to constructing a subset of the data.

Our first subset is a \emph{uniformly random sample} of $50{,}000$ sequences, drawn from the training dataset without repetition.
While a uniform sample is useful to estimate the absolute amount of memorization in a model, it is poorly suited for studying how memorization scales with data properties that are \emph{not} uniformly represented in the training set. 
For example, prior work has identified that \emph{data duplication} (i.e., how often the same sequence is repeated either exactly or approximately) is an important factor for memorization.
Yet, because the frequency of training data duplication decays extremely quickly \citep{2021dedup}, a uniformly random sample of $50{,}000$ sequences
(accounting for $\le 0.02\%$ of the dataset)
is unlikely to contain \emph{any} signal that would allow us to
accurately measure the tail of this repeated data distribution. A similar concern arises for measuring how memorization scales with prompt length, since very long sentences account for only a small fraction of the training set.

Therefore, our second subset is a random sample \emph{normalized} by both sequence lengths and duplication counts, which allows us to accurately measure memorization of large language models in the worst-case, on highly duplicated data with long prompts.
For each sequence length $\ell \in \{50,100,150,\dots,500\}$,
and integer $n$,
we select $1{,}000$ sequences of length $\ell$ that are contained in the training dataset between $2^{n/4}$ and $2^{(n+1)/4}$ times.
We do this until we reach an $n$ for which $1{,}000$ sequences are not available.
This gives us $1{,}000$ sequences that repeat between $6$ and $8$ times ($\approx 2^{11/4}$ and $\approx 2^{12/4}$) and also $1{,}000$ sequences that repeat between $724$ and $861$ times ($\approx 2^{38/4}$ and $\approx 2^{39/4}$).
This biased sampling allows us to more accurately measure memorization as a function of a sample's duplication factor and prompt length, without querying the entire dataset.
Note that constructing this duplicate-normalized data subset requires some work,
as efficiently identifying duplicate substrings in an $800$GB training dataset is
computationally challenging.
We make use of the suffix array construction from \citet{2021dedup} (see Appendix).

For each length from $50$ to $500$ tokens, we collect $50{,}000$ examples duplicated varying numbers
of times,
totaling roughly $500{,}000$ sequences.
For each sequence of length $\ell$, we prompt the model with the first $\ell-50$ tokens and report the sequence
as ``extractable'' if the model exactly emits the next $50$ token suffix of this sequence.
Fifty tokens corresponds to an average of 127 characters or 25 words
 in the GPT-Neo training set, well over the length of a typical English sentence.
Finally, we compute the average probability that a sequence is extractable by averaging over all lengths $\ell$.

%

\begin{figure*}
\hspace{-2em}
    \centering
    \begin{subfigure}[b]{0.37\textwidth}
        \includegraphics[height=12.2em]{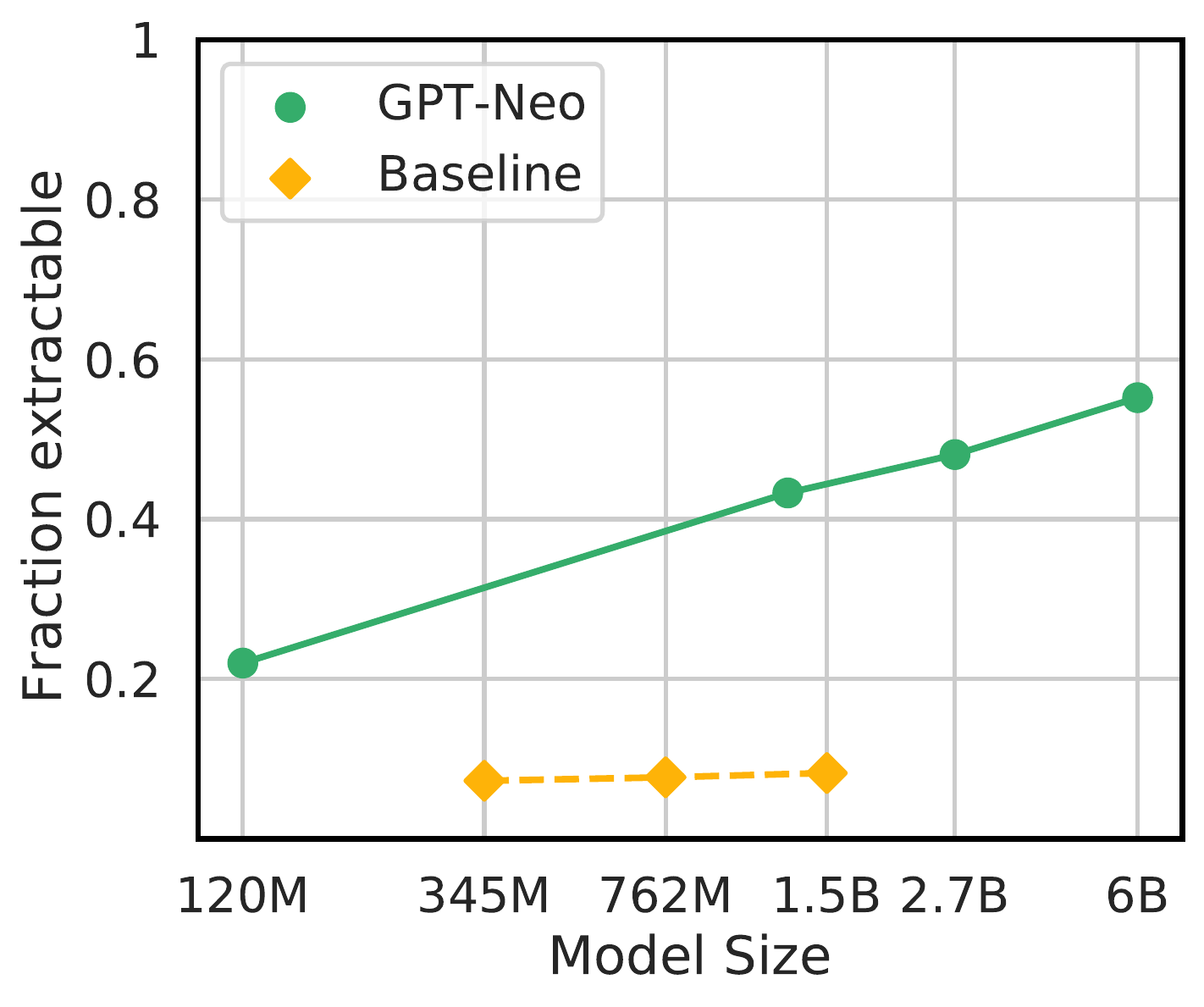} 
        \caption{}
        \label{fig:main-res-size}
    \end{subfigure}
    \begin{subfigure}[b]{0.32\textwidth}
        \includegraphics[height=12em]{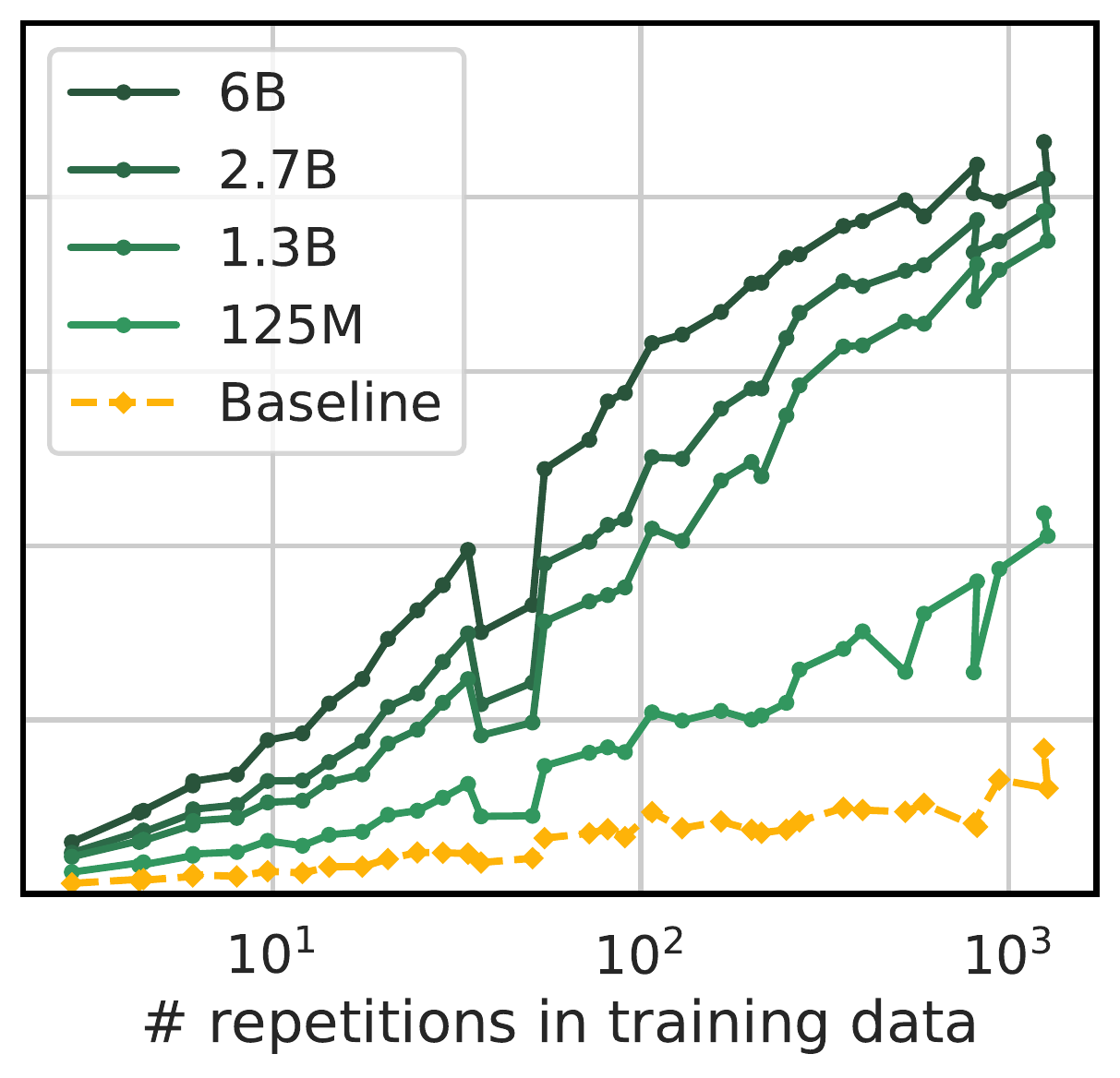} 
        \caption{}
        \label{fig:main-res-dups}
    \end{subfigure}
    \begin{subfigure}[b]{0.33\textwidth}
        \includegraphics[height=12em]{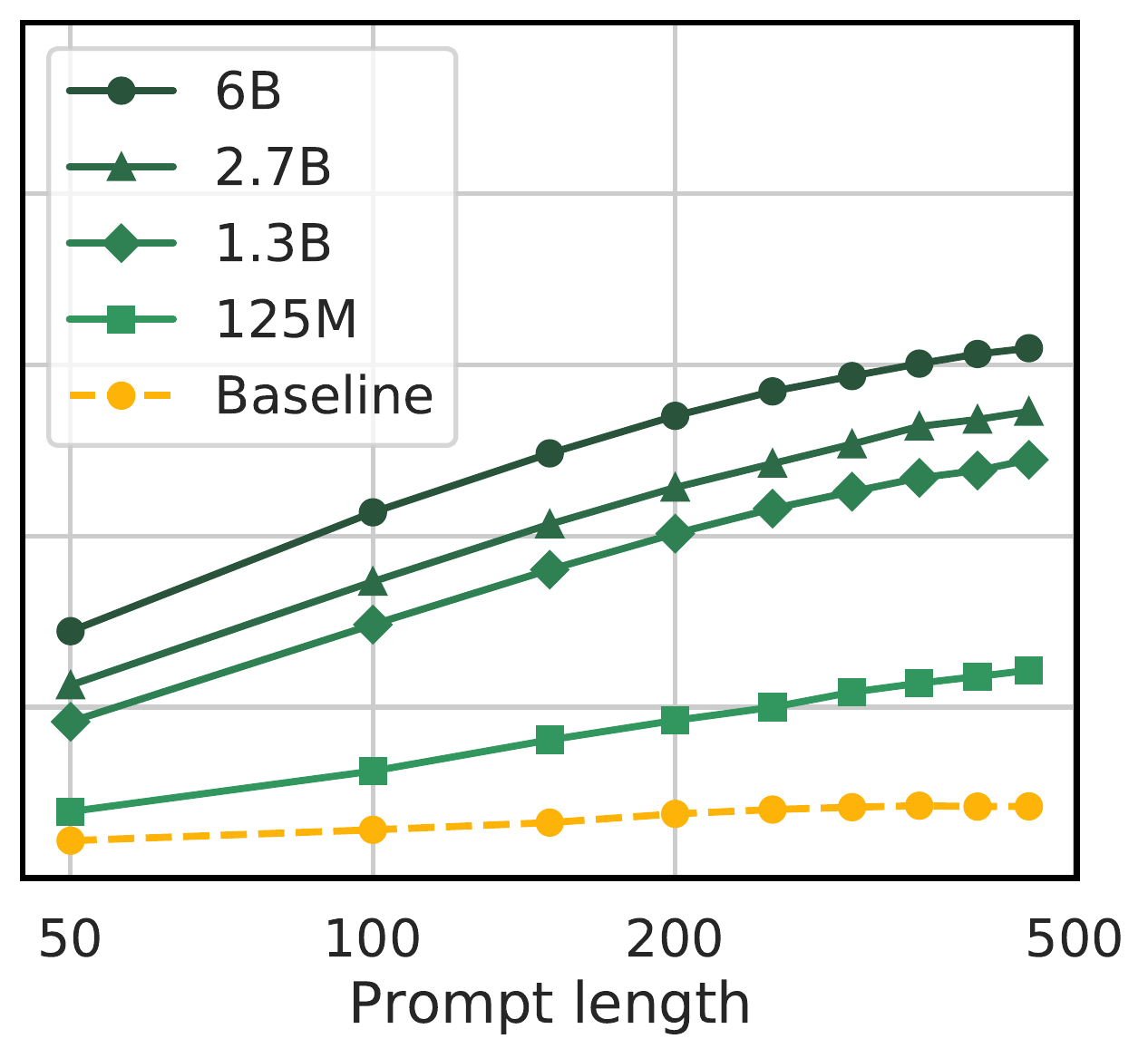} 
        \caption{}
        \label{fig:main-res-context}
    \end{subfigure}
    \vskip-5pt
    \caption{
    We prompt various sizes of GPT-Neo models (green) with data sampled from their training set---The Pile, and normalized by sequence lengths and duplication counts.
    As a baseline (yellow), we also prompt the GPT-2 family of models with the same Pile-derived prompts, even though these models were trained on WebText, a different training dataset.
    \textbf{(a)}
    Larger models memorize a larger fraction of their training dataset, following a log-linear relationship.
    This is not just a result of better generalization, as shown by the lack of growth for the GPT-2 baseline models.
    \textbf{(b)}
    Examples that are repeated more often in the training set are more likely to be extractable,
    again following a log-linear trend (baseline is GPT-2 XL).
    \ \textbf{(c)}
    As the number of tokens of context available increases, so does our ability to extract memorized text (baseine is GPT-2 XL).
    }
    \label{fig:main-res}
\end{figure*}

\section{Experiments}

We primarily study the GPT-Neo model family~\citep{gpt-neo,gpt-j} trained on the Pile dataset~\citep{gao2020pile}.
The GPT-Neo models are causal language models 
trained with the objective of predicting the next token in a sequence given the previous ones.
They come in four sizes: $125$ million, $1.3$ billion, $2.7$ billion and $6$ billion parameters.\footnote{As of February 2022, there is also a $20$ billion parameter variant. Unfortunately this model uses a different training setup and tokenizer making it difficult to apply here.} 
The Pile is a dataset of $825$GB of text collected from various sources (e.g., books, Web scrapes, open source code).
Prior to the recent release of OPT~\citep{zhang2022opt}, the GPT-Neo models were the largest language models available for public download, and The Pile is the largest public text dataset available.

\subsection{Bigger Models Memorize More}
\label{sec:modelsize}
We begin by considering the impact of model size on memorization,
expanding on prior studies which qualitatively established a relationship between the size of GPT-2 models and their
ability to memorize $<$30 URLs \citep{carlini2020extracting}.
%
In contrast, we study \emph{a million} model generations in order to describe
how model scale relates to memorization.

\minisection{Results.}
We first study our biased random data sample normalized by duplication count and sequence lengths. The results of this experiment are given in Figure~\ref{fig:main-res-size}.
The y-axis reports the fraction of generations which exactly reproduce the true suffix for their prompt, averaged over all prompt and sequence lengths in our evaluation set. Because our biased sampling over-represents duplicated strings, the \emph{absolute} degree of memorization in Figure~\ref{fig:main-res-size} is not particularly important here---rather, we are interested in how memorization varies with scale.\footnote{We repeat this experiment for a uniformly random subset of the data in Figure~\ref{fig:other_approaches_randomsize}.}
We find that larger models memorize significantly more than smaller models do, with \emph{a near-perfect log-linear fit} ($R^2$ of 99.8\%): a ten fold increase in model size corresponds to an increase in memorization of 19 percentage points.

To confirm that larger models are indeed \emph{memorizing} more data, and not simply \emph{generalizing} better, we repeat the analysis with the GPT-2 model family as a baseline. 
The GPT-2 models are similarly sized, and also trained on Internet-scraped data.
If our ``larger models memorize more'' result was due to the predictive strength of larger models, and not the memorization of specific training data, we would expect a similar relationship between comparably sized GPT-2 models trained on similar data. 
Put differently, this baseline allows to establish
what fraction of the training data is sufficiently ``easy'' that any language model can correctly
predict the 50-token suffix, even if the example has not been seen during training.
For example, a language model trained on multiple examples of number sequences can likely correctly complete some other unseen number sequences.
%
%
We find that GPT-2 correctly completes approximately $6\%$ of the examples in our evaluation set, compared to $40\%$ for the similarly sized 1.3B parameter GPT-Neo model. 
A qualitative analysis (see examples in Appendix Figure~\ref{fig:egs-mem-by-gpt2}) suggests that examples ``memorized'' by GPT-2 are largely uninteresting
sequences (e.g., number sequences, repetitions of the same few tokens, or common phrases).
Therefore, we conclude that larger models have a higher fraction of extractable training data because they have actually memorized the data; it is not simply that the larger models are more accurate.

\subsection{Repeated Strings are Memorized More}
\label{sec:duplicates}
Prior work provides preliminary evidence that memorization in language models increases with the number of times sequences are repeated in the training set \citep{carlini2020extracting, 2021dedup}.
We expand on this observation and quantitatively measure the effect of data duplication on memorization.
Using our duplication-normalized data sample, we measure the fraction of sequences which are extractable, for buckets of sequences duplicated between 2 and 900 times.
Each bucket consists of $1{,}000$ distinct sentences, and we compute the average amount of memorization for each bucket.

\minisection{Results.}
Figure~\ref{fig:main-res-dups} shows our results, aggregated over all sequence lengths.
We observe a clear log-linear trend in memorization. While models rarely regurgitate strings that are repeated only a few times, this probability increases severely for highly duplicated strings.
%
The small memorization values at low numbers of repetitions corroborates the positive impact of training dataset \emph{deduplication} on memorization observed by \citet{2021dedup}.
However, we find that memorization does still happen, even with just a few duplicates---thus, deduplication will not perfectly prevent leakage.
While this relationship is perhaps obvious, and has been corroborated for specific training examples in prior work \citep{carlini2019secret, carlini2020extracting}, our results show that it holds \emph{across the entire training set}.

\subsection{Longer Context Discovers More Memorization}
\label{sec:context}
The previous two questions evaluated how data collection and model training decisions impact the leakage of a model's training data when it is provided a fixed number of tokens from a sequence as context.
As a result, those experiments suggest particular actions that could be taken to mitigate memorization (by reducing model size, or limiting the number of duplicate examples).

However, even when the model is fixed, it is possible to vary the amount of extractable training data by controlling the length of the prefix passed to the model. 
By studying how the number of tokens of context impacts extractability, we demonstrate the difficulty of \emph{discovering} memorization---language models may only exhibit their memorization under favorable conditions.



\minisection{Results.}
In Figure~\ref{fig:main-res-context}, we observe that the fraction of extractable sequences increases log-linearly with the number of tokens of context. For example, 33\% of training sequences in our evaluation set are extractable from the 6B model at 50 tokens of context, compared to 65\% with $450$ tokens of context.
We call this the \textbf{discoverability phenomenon}:
some memorization only becomes apparent under certain conditions, such as when the model is prompted with a sufficiently long context.

%

The discoverability phenomenon may seem natural: conditioning a model on $100$ tokens of context is more specific than conditioning the model on $50$ tokens of context, and it is natural that the model would estimate the probability of the training data as higher in this situation. 
However, the result is that some strings are ``hidden'' in the model and require more knowledge than others to be extractable.

From one point of view, it is good that some memorization is difficult to discover.
This makes it harder for attackers to perform training data extraction attacks \citep{carlini2020extracting},
or otherwise exploit memorization.
Indeed, if an exact $100$ token prompt is required to make the model output a given string, then, in practice, an adversary will likely be unable to perform the attack.
The difficulty in discovering memorization also reduces the likelihood of \emph{non-adversarial} training data regurgitation.
For example, the GitHub Copilot model~\citep{chen2021evaluating} reportedly rarely emits memorized code in benign situations, and most memorization occurs only when the model has been prompted with long code excerpts that are very similar to the training data~\citep{ziegler2021copilot}.
Practitioners building language generation APIs could (until stronger attacks are developed) significantly reduce extraction risk by restricting the maximum prompt length available to users.

Viewed differently, however, the difficulty of discovering memorization
can also harm our ability to audit privacy in machine learning models.
Because provably-correct approaches for privacy-preserving training of machine learning models are applied only rarely in practice~\citep{abadi2016deep, thakkar2020understanding, ramaswamy2020training},
it is common to attempt post-hoc \emph{privacy auditing}~\citep{jayaraman2019evaluating, jagielski2020auditing, nasr2021adversary}.
Our results suggest that correctly auditing large language models likely requires prompting the model with training data, as there are no known techniques to identify the tail of memorized data without conditioning the model with a large context.
Improving upon this limitation is an interesting problem for future work.

\begin{figure}
    \centering
    \begin{subfigure}[a]{0.34\textwidth}
        \includegraphics[width=\textwidth]{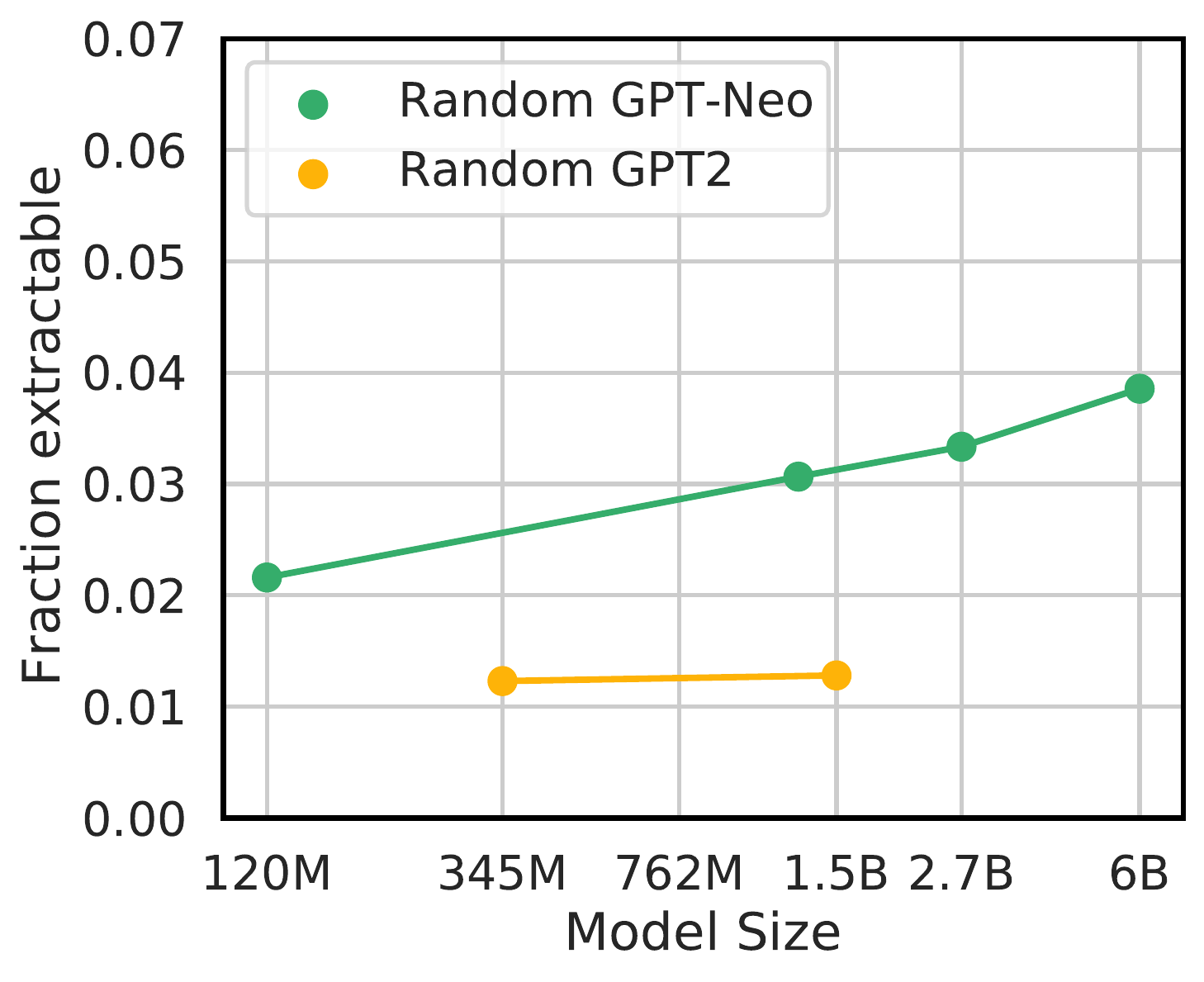} 
        \caption{}
        \label{fig:other_approaches_randomsize}
    \end{subfigure}
    \hfill
    \begin{subfigure}[a]{0.34\textwidth}
        \includegraphics[width=\textwidth]{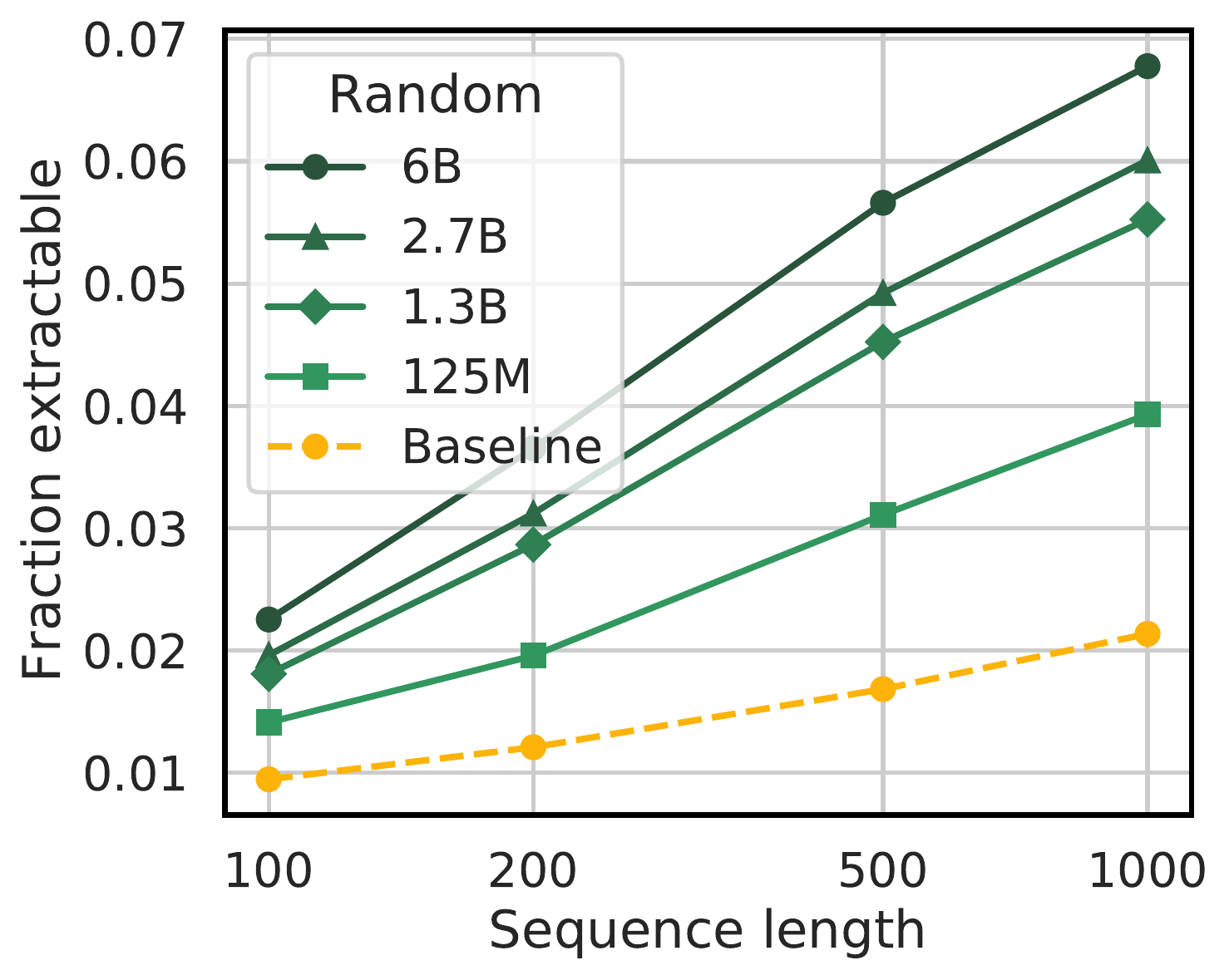} 
        \caption{}
        \label{fig:other_approaches_randomlength}
    \end{subfigure}
    \hfill
    \begin{subfigure}[a]{0.3\textwidth}
        \includegraphics[width=\textwidth]{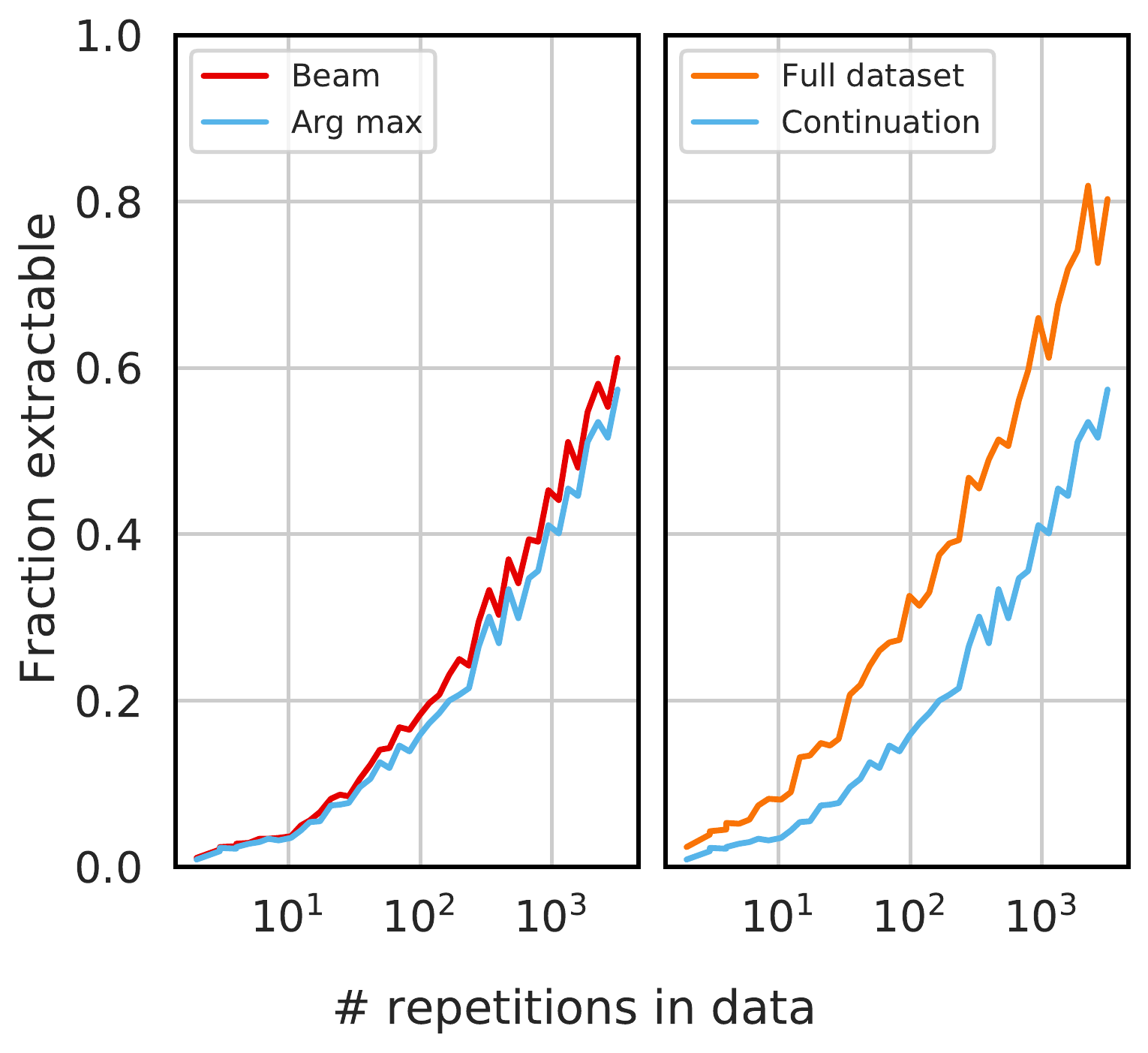} 
        \caption{}
        \label{fig:other_approaches_search}
    \end{subfigure}
    
    \vskip-5pt
    
    \caption{
    \textbf{(a)} Fraction of sequences extracted as a function of model scale where we sample uniformly from the training set. 
    \textbf{(b)} Fraction of sequences extracted as we vary the length of the prompt. For each sequence length $n$, $n$-50 tokens are used as the prefix, and we check for extraction of the remaining 50 tokens.
    \textbf{(c-left)} Using beam search with $b$=100 slightly increases the data extracted. \textbf{(c-right)} We observe considerably more memorization when checking whether the generated sequence occurs anywhere in the entire training set (Section \ref{section:other_approaches}). However, this approach is very computationally expensive so we do not use it for our other experiments.}
    \label{fig:other_approaches}
\end{figure}

\subsection{Alternate Experimental Settings}
\label{sec:alt_settings}
In this section, we briefly review other strategies that we could have used to quantify memorization.
%

\minisection{Random dataset sampling.}
The majority of this paper uses subsets of the training data that were explicitly sampled according to training data duplication frequency.
Now, we consider how our results would differ if we chose a
truly random subset of the training data, where each sequence is sampled uniformly,
instead of sampling a duplicate-normalized dataset.
Specifically, we randomly sample $100{,}000$ sequences of varying lengths from The Pile dataset, then prompt the model and test for memorization as before (more details in Appendix~\ref{section:other_approaches}).
%

Figure~\ref{fig:other_approaches_randomsize} and Figure~\ref{fig:other_approaches_randomlength}
present the results.
We observe similar qualitative trends with model scale and context length as in Figure~\ref{fig:main-res}.
Larger models memorize more training examples than smaller models---and much more than the GPT-2 models
that were not trained on The Pile.
Similarly, providing more context to a model increases
the likelihood we discover memorization.
We can extract the last 50 tokens of
a length-1000 sequence with $7\%$ probability
for the largest GPT-J 6B model, compared to $4\%$ probability 
for the smallest $125$M GPT-Neo model. (And both
of these are much larger than the $2\%$ probability 
of extraction for the $1.5$B parameter GPT2-XL model.)
These results, taken together, allow us to estimate a lower bound that
there is at least $1\%$ of The Pile dataset that is extractable
by the 6B GPT-J model, but not by GPT-2 XL.

\minisection{Alternate decoding strategies.}
We have defined memorization as a model's ability to generate the true continuation when choosing the \emph{most likely} token at every step of decoding.
Yet, this greedy decoding strategy does not produce the overall most likely sequence.
Many language model applications use other decoding strategies, such as beam search 
to find the generation with highest likelihood.
To understand how our choice of decoding strategy affects the amount of memorization we measure, we compare greedy decoding with beam search in Figure \ref{fig:other_approaches}(c).
We find that using beam search with 100 beams results in marginally more extracted memorization.
The difference in extractable memorization is just under 2 percentage points on average, with a maximum of 5.6\%. 
Interestingly, beam search and greedy decoding generated the same output 45\% of the time.

The most common decoding strategy employed by modern LMs is \emph{random sampling}, where the next token is selected at random according to a probability distribution derived from the model's predictions.
\citet{mccoy2021raven} found that random sampling resulted in generated text with a greater number of novel $n$-grams.
Since the goal of our study is to maximize discoverability---an antithetical goal to maximizing linguistic novelty---we do not present experiments that use random sampling.

\begin{figure*}
    \centering
    \includegraphics[width=\linewidth]{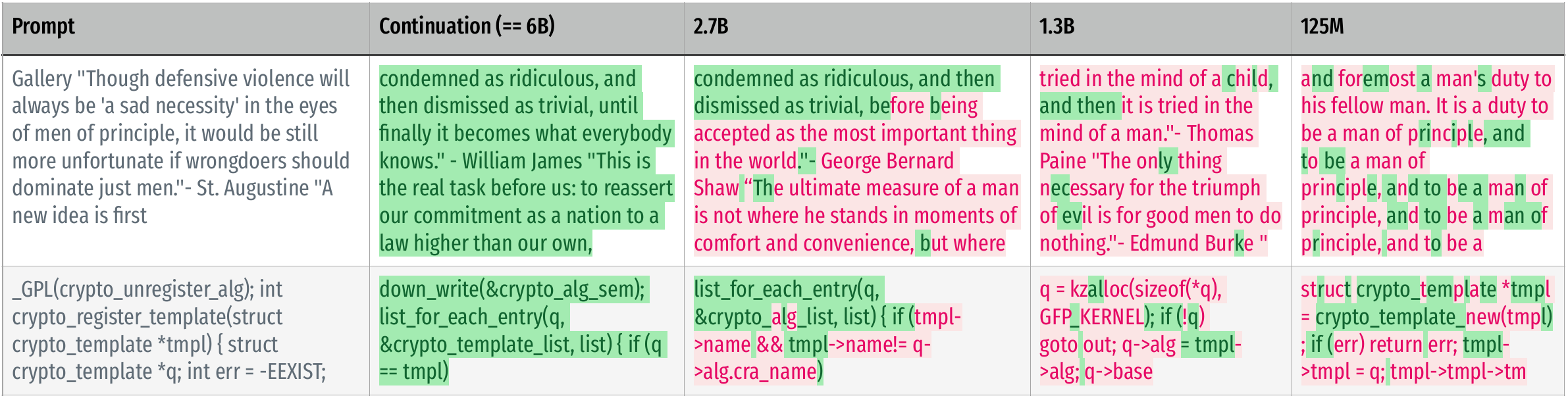}
    \caption{Text examples that are memorized by the 6B model, but not by smaller models. Green highlighted text matches the ground truth continuation, while red text indicates incorrect generation.}
    \label{fig:egs-mem-by-6b-body}
    \vspace{-0.5em}

\end{figure*}

\minisection{Alternate definition of extractability.}
Our main experiments report a sequence as ``extractable'' if the model's generation is identical to the true suffix of the considered training example.
However it is possible this suffix is still present (elsewhere) in the dataset.
We now consider a loose lower bound on memorization that considers a sequence memorized if the generation $[p || f(p)]$ from a prompt $p$ is contained \emph{anywhere} in the training dataset.
Searching within the entire dataset finds more memorized content than comparing with the ground truth (Figure \ref{fig:other_approaches_search}). 
For examples at 100 repetitions, $32.6\%$ of outputs are contained somewhere in the dataset but just $15.8\%$ match the ground truth continuation.

\subsection{Qualitative Examples of Memorization}
In Figure~\ref{fig:egs-mem-by-6b-body}, we present qualitative examples that are only memorized by the largest (6B) model, but not the smaller ones. We highlight some interesting patterns in these sequences: while the generations from the smaller models do not match the training data, they are generally thematically-relevant and locally consistent. However, a closer inspection reveals that those generations are only syntactically sound, but semantically incorrect.
Appendix  Figure~\ref{fig:egs-mem-by-all} shows further examples of sequences that are memorized by \emph{all} the models. We found most of these universally-memorized sequences to be ``unconventional'' texts such as code snippets or highly duplicated texts such as open source licenses. 
Figure~\ref{fig:egs-mem-but-not-rep} shows sequences which are memorized by the 6B parameter model despite being infrequent in the training set.
These tend to be easily completed text--
Figure~\ref{fig:egs-rep-but-not-mem} shows sequences which are repeated thousands of times but are surprisingly not memorized by the 6B parameter model.
Many of these are mostly correctly completed, only differing on semantically unimportant characters.

\section{Replication Study}

The above analysis provides evidence that memorization scales log-linearly with model size, data duplicates, and context length.
We now replicate this analysis for other language models trained with different datasets and training objectives, namely:
(1) the T5 family of models trained on the C4 dataset~\citep{t52020},
(2) models from \citet{2021dedup}, trained on a deduplicated version of C4, and
(3) the OPT family of models \citep{zhang2022opt}, also trained on the Pile.
%
We expected our results to cleanly generalize across settings, and this is indeed true for model scale. Yet, the situation is more complicated when considering data duplication, due to training set idiosyncrasies.

\subsection{T5 Masked Language Modeling}

\begin{figure*}[t]

    \centering
    \begin{subfigure}[b]{0.32\textwidth}
        \includegraphics[height=10.5em]{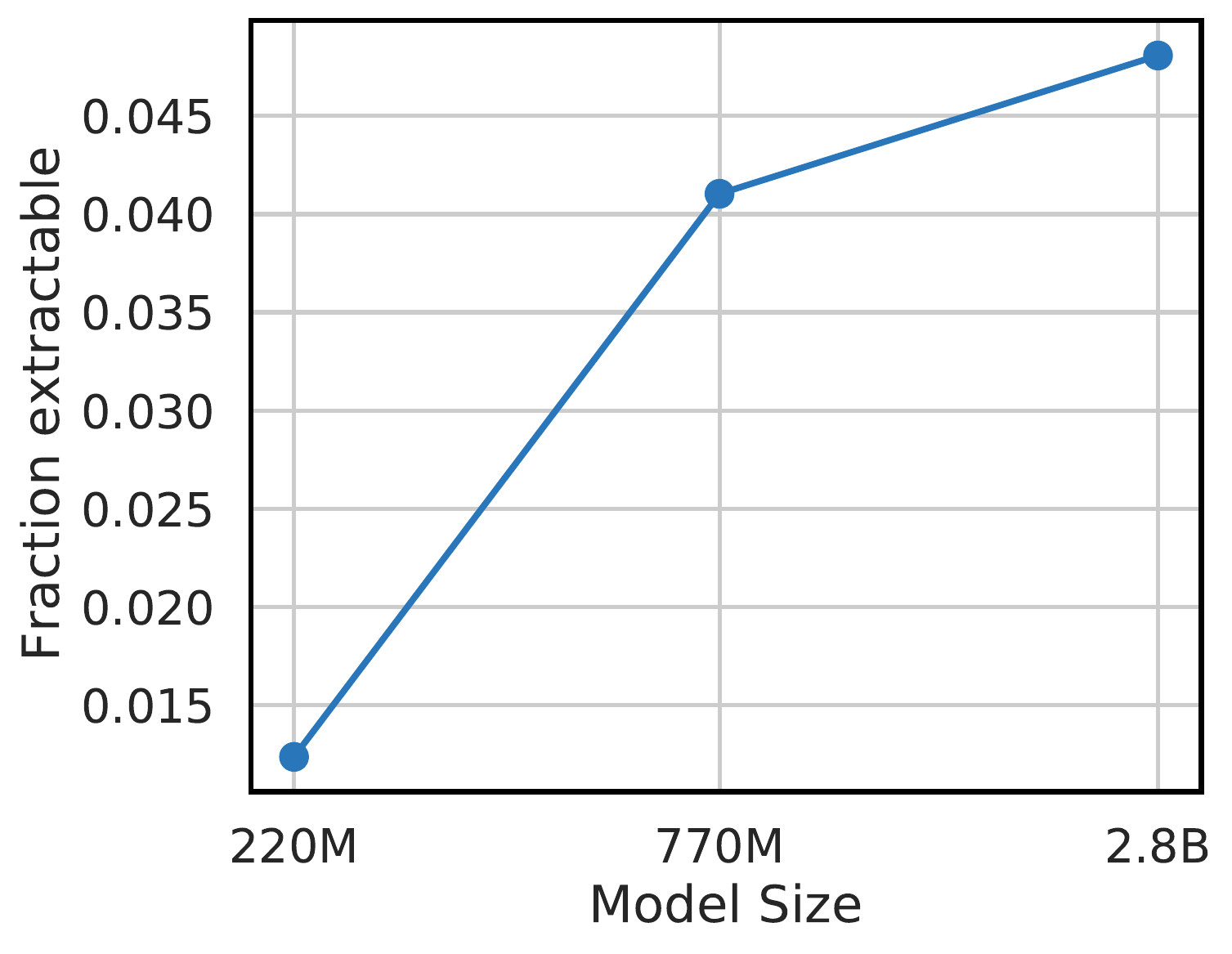}
        \caption{}
        \label{fig:other-models-size}
    \end{subfigure}
    \hfill
    \begin{subfigure}[b]{0.32\textwidth}
        \includegraphics[height=10.5em]{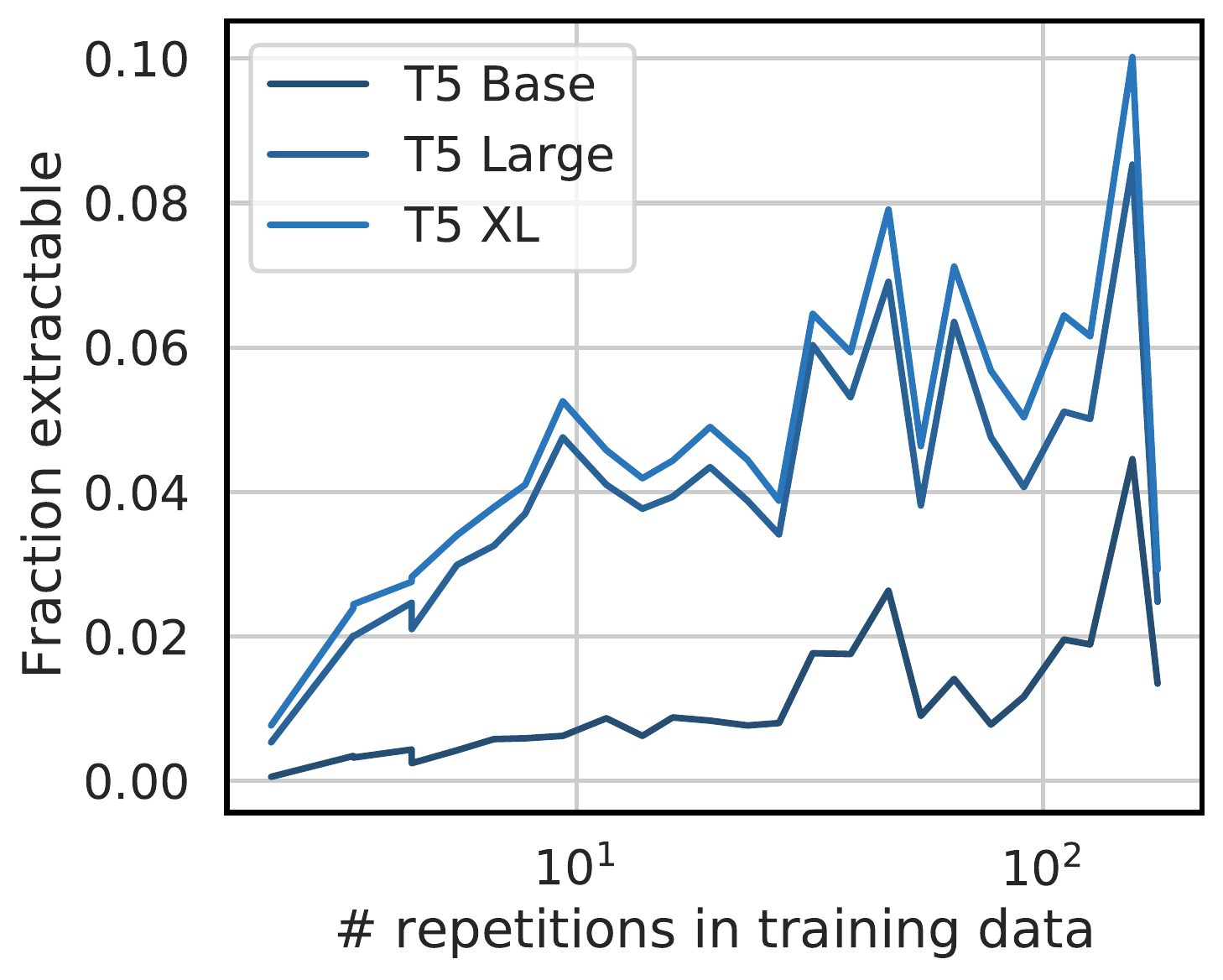}
        \caption{}
        \label{fig:other-models-dups}
    \end{subfigure}
    \hfill
    \begin{subfigure}[b]{0.32\textwidth}
        \includegraphics[height=10.5em]{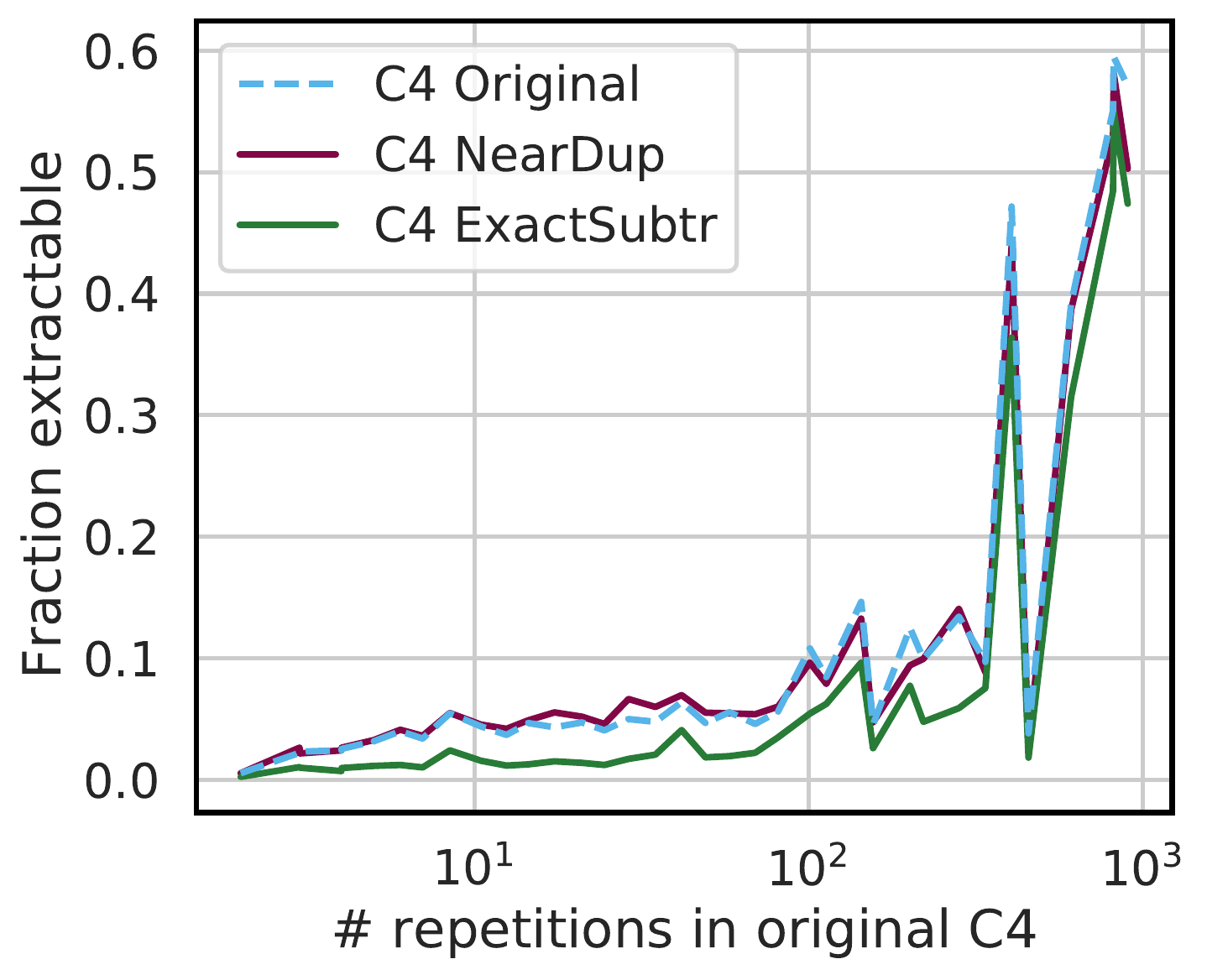}        \caption{}
        \label{fig:other-models-dedup}
    \end{subfigure}

    \vskip-5pt
    \caption{
    \textbf{(a)}
    Masked language model objective: Larger models have a higher fraction of sequences extractable on T5.
    \textbf{(b)}
    Masked language model objective: Relationship between number of repetitions and extractable tokens on T5.
    \textbf{(c)}
    Causal language model objective: Relationship between number of repetitions and memorization on language models trained with deduplicated data.
    }
    \label{fig:other-models}
\end{figure*}
\minisection{Model and dataset.}
The T5 v1.1 models are masked encoder-decoder models trained to reproduce randomly deleted spans from an input sequence. The models vary in size from  77M to 11B billion parameters, and are trained on C4---a 806 GB curated version of English web pages from the Common Crawl.
The largest T5 model (11B parameters) is the largest publicly available masked language model. T5 models are thus good candidates for studying how memorization scales with model size.

We must first define what is meant by ``extractable data''
for the masked language modeling task.
T5 models are trained by removing a random $15\%$ of tokens from each training sequence (i.i.d), and the model must then
``fill in the blanks'' to restore the tokens that were dropped from the input.
%
%
As a result of this different training objective,
Definition~\ref{def:extractable} is not directly applicable: the model does not operate on a \emph{prefix} and output a \emph{suffix}.
We instead call a sequence memorized if the model \emph{perfectly} solves the masked language modeling task on that sequence.
For example, we call a 200-token sequence memorized if the model can
use the 170 ($=200 \cdot 0.85)$) tokens of context to perfectly predict the remaining 30 tokens ($=200 \cdot 0.15$).
Because this token-dropping procedure is stochastic, it is possible that one set of dropped tokens might yield
an output of ``memorized'' and another might not.
For simplicity, we inspect only one set of masked tokens per sequence; because we are already averaging over
$50{,}000$ sequences this additional randomness does not harm the results of our analysis.

\minisection{Results.}
In Figure~\ref{fig:other-models-size}, we reproduce the model scaling effect (from Figure~\ref{fig:main-res-size}) for T5 models.
Larger models similarly have an increased ability to perfectly solve the masked prediction task.
Surprisingly, while a scaling trend does hold here as well, the absolute memorization in masked models is an order of magnitude lower than for comparably
sized causal language models.
For example, the $3$B parameter T5-XL model memorizes $3.5\%$ of sequences repeated 100 times, whereas the GPT-Neo 2.7B model memorizes $53.6\%$ of sequences repeated 100 times (with 150 tokens of context). 

Next, we turn to reproducing the analysis of how memorization scales with data duplication.
The situation here becomes significantly less clear.
As shown in Figure~\ref{fig:other-models-dups}, sequences duplicated more often tend to be easier to memorize, but there is no monotonic scaling relationship.
Compared to the case of the GPT-Neo models trained on The Pile, the relation between data duplication counts and memorization for T5 models trained on C4 exhibits large variance.
This variance is \emph{statistically significant}:
sequences repeated 159 to 196 times are memorized with probability less than 5.1\% with $99.7\%$ confidence (three standard deviations from the mean),
however sequences repeated 138 to 158 times (that is, \emph{less often}) are memorized with probability at least 6.2\% (also with $99.7\%$ confidence).
That is, for some reason, sequences that occur $\sim$140 times are \emph{more likely to be memorized, despite occurring less often}, even if we assume a three-sigma error in both measurements simultaneously.

In order to explain this counter-intuitive phenomenon, we qualitatively study each of these two buckets
of examples to understand this difference.
We find that most of the duplicate examples repeated 138-158 times consist mainly of whitespace tokens. These sequences are thus much easier to predict correctly than other sequences, even if they are repeated more often.
This effect, to a lesser extent, can be found in other buckets which contain many approximately near duplicates.

\subsection{Language Models Trained on Deduplicated Data}
\minisection{Model and dataset.}
The models used in \citet{2021dedup} are 1.5B parameter causal language models.
This model family consists of one model trained on C4 (the same dataset as T5), one model trained on a version of C4 that was deduplicated by removing all documents which were near-duplicates of other documents, and one model trained on a version of C4 that was deduplicated by deleting any string of length-50 tokens that occurred more than once.
\citet{2021dedup} found that both types of deduplication reduced the likelihood of memorization.

\minisection{Results.}
We were most interested in whether models trained on deduplicated data would still exhibit increased memorization of examples which were repeated frequently in the original, non-deduplicated C4 dataset (e.g., because the deduplication missed some near-duplicates).
%
Figure~\ref{fig:other-models-dedup} plots the fraction of sequences memorized by these three models.
We draw two interesting conclusions from this data.

First, we confirm that models trained on deduplicated datasets memorize less data than models trained without deduplication.
For example, for sequences repeated below 35 times, the exact deduplicated model memorizes an average of 1.2\% of sequences, compared to 3.6\% without deduplication, a statistically significant ($p<10^{-15}$) decrease by a factor of 3$\times$. 
Second, while deduplication does help for sequences repeated up to $\sim$100 times,
it does not help for sequences repeated more often!
The extractability of examples repeated at least 408 times is statistically significantly higher than any other number of repeats before this.  
%
We hypothesize that this is due to the fact that any deduplication strategy is necessarily
imperfect in order to efficiently scale to hundreds of gigabytes of training data.
%
Thus, while it may be possible to remove \emph{most} instances of duplicate data, different and valid definitions of duplicates can mean deduplication is not exhaustive.

\subsection{Language Models Trained on a Modified Version of the Pile}

\paragraph{Model and dataset.}
We finally study the OPT family of models \citep{zhang2022opt}, that vary from 125 million to
175 billion parameters.%
\footnote{We were unable to access the 175 billion parameter model; we run OPT models up to 66 billion parameters.}
These models were trained on a $800$GB dataset that overlaps with The Pile but is not identical and contains data from many new sources, while also removing some data from the Pile.
This dataset was also deduplicated prior to training, and so we do not expect to see duplicate sequences
memorized (much) more than sequences repeated only a few times.

\paragraph{Results.}
Overall, we find that while there are nearly identical scaling trends to those we found for
GPT-Neo's model family, 
the effect size is orders-of-magnitude smaller (figure \ref{fig:opt}).
Even the 66 billion parameter model memorizes a smaller fraction of The Pile than
the smallest 125 million parameter GPT Neo model.
This suggests two possible conclusions:
(a) careful data curation and training can mitigate memorization, or
(b) even slight shifts in data distribution can significantly alter what content gets memorized.
%
Without direct access to the original training dataset, we can not distinguish between these
two conclusions and hope future work will be able to resolve this question.

\section{Conclusion}

%
Our paper presents the first comprehensive quantitative analysis of memorization in
large language models, by re-processing the training set to find memorized data.
Our work has two broad conclusions.

For the study of generalization, we have shown that while current LMs do accurately model the statistics of their training data, this need not imply that they faithfully model the desired \emph{underlying} data distribution.
In particular, when the training data distribution is skewed (e.g., by containing many duplicates of some sequences) larger models are likely to learn these unintended dataset peculiarities.
It is therefore important to carefully analyze the datasets used to train ever larger models, as future (larger) models are likely to remember even more training details than current (smaller) models.

For the study of privacy, our work indicates that current large language models memorize a significant fraction of their training datasets.
Memorization scales log-linear with model size---by doubling the number of parameters in a model we can extract a significantly larger fraction of the dataset.
Given that current state-of-the-art models contain more than 200$\times$ as many parameters as the largest 6B parameter model we analyze, it is likely that these even larger models memorize many sequences that are repeated just a handful of times.
At the same time, we have shown that this memorization is often hard to discover, and for an attack to actually extract this data it will be necessary to develop qualitatively new attack strategies.
Fortunately, it appears that (for the comparatively small models we study) training data inserted just once is rarely memorized, and so deduplicating training datasets~\citep{2021dedup} is likely a practical technique to mitigate the harms of memorization.

\bibliographystyle{plainnat}
\bibliography{refs}


\clearpage\appendix

\section{Implementation Details for Dataset Creation}
\label{sec:suffixarray}

Intuitively speaking, it is straightforward to construct a dataset containing specifiable proportions of documents at various frequencies.
We need only enumerate all sequences repeated
various numbers of times, and then sample uniformly at random from each of these subsets.
However in practice this is difficult to do, given the scale of these datasets: even asking
the question ``how many times is this sequence present in the training dataset'' requires
linear work for each query, and so repeating this thousands of times for an $800$GB dataset
would be infeasible.

To do this efficiently, 
we build on the work of \citet{2021dedup} and construct a suffix array over the training dataset.
Such a data structure allows efficient queries to enumerate all sequences of length $k$ that are repeated
between $N$ and $M$ times for any $N,M$. This can be accomplished by a linear scan of the suffix array. 
As notation, write $i$ as the pointer into the dataset at a certain position $j$ of the suffix array (i.e., $A[j] = i)$, $i'$ as the index at position $j+N$ (so that $A[j+N]=i'$), 
and $i''$ as the index at position $j+M$ (so that $A[j+M] = i''$.
Then, if $D[i:i+k] = D[i':i'+k]$ but $D[i:i+k] \ne D[i'':i''+k]$,
the sequence $D[k:i+k]$ is guaranteed to appear between $N$ and $M$ times in the dataset.
As a result, we can scan linearly through the suffix array and enumerate all values of j $j$ to efficiently find all potential sequences repeated between N and M times. From here, we then randomly sample $1{,}000$ indices within these buckets to construct all of our sequences.

\section{Longer Documents Are Not Easier to Memorize than Shorter Documents}
\begin{figure}[h]
    \centering
    \includegraphics[width=0.5\textwidth]{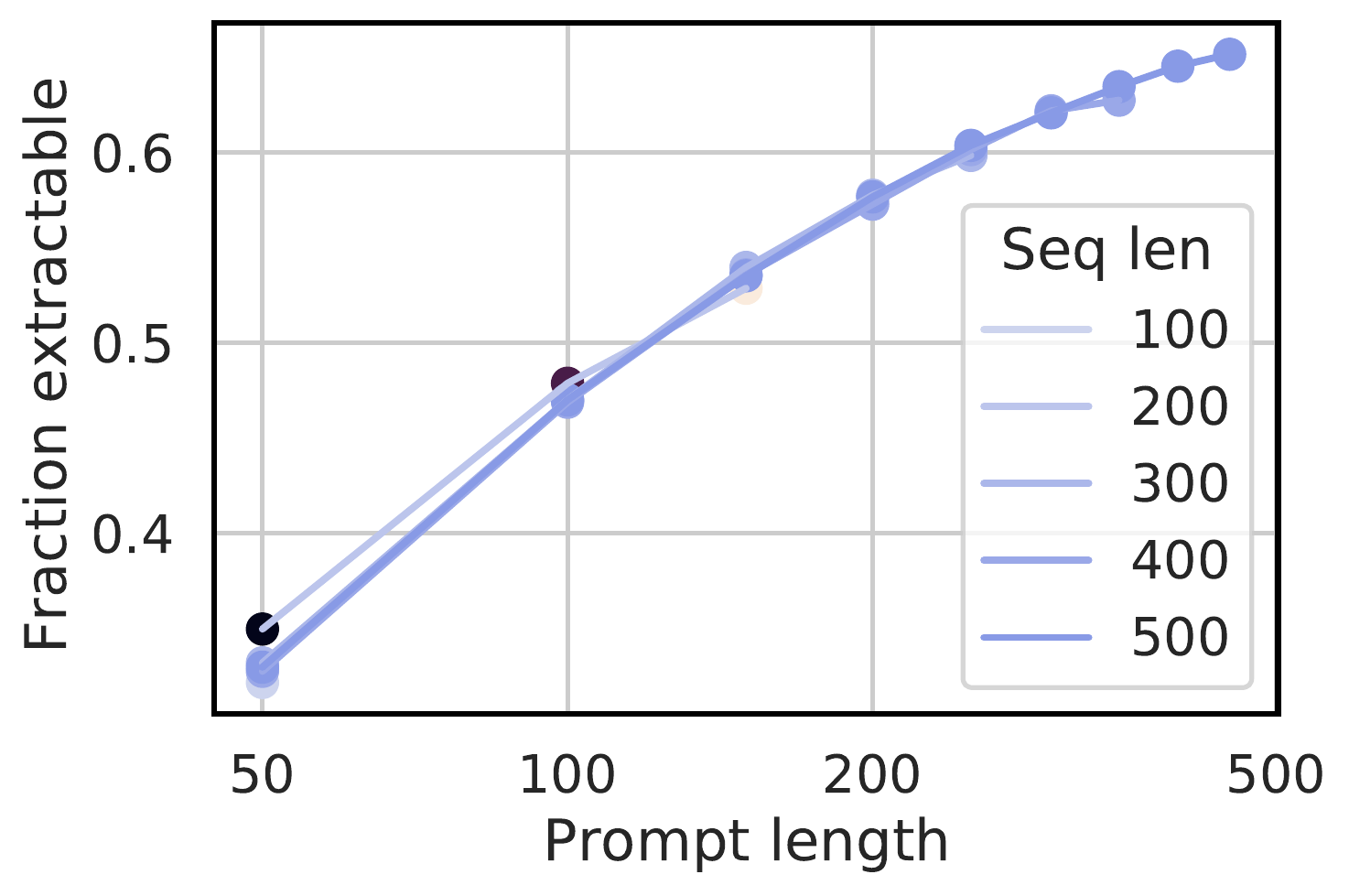}
    \caption{
    Longer sequences are not easier to extract.
    We compute the probability that an adversary can extract a sequence as a function
    of the number of tokens of context available, when varying the length of the sequences.
    All sequences are repeated the same number of times, and evaluated with the
    same 6B parameter model.
    Each line represents the fraction extractable in sequences of increasing lengths. 
    Because all lines nearly perfectly overlap, longer sequences are not fundamentally ``easier'' to extract than shorter sequences.
    }
    \label{fig:context-seq}
\end{figure}

Intuitively, one might think that longer sequences are more likely in the tail of the distribution, and if the model is trained to a low perplexity, then the tail of the distribution may be more likely to be memorized.
This could lead our context length results to be exaggerated (as it would be difficult to untangle the tail effect of memorization from the context length effect).
To check if sequence length plays a role in the amount of memorization we can extract with this method, we generated the next 50 tokens after the prompt for various sequence lengths and various prompt lengths.
Figure \ref{fig:context-seq} shows the fraction of extractable tokens in the next 50 tokens after the prompt. 
Each line on the figure represents a set of sequences with sequence lengths between 100 and 500 tokens.
For each sequence length, we looked at prompt lengths from 50 tokens to $(\text{sequence length} - 50)$ tokens.
We do not see significant differences between the fraction of extractable tokens with varying prompt lengths across various sequence lengths.

\section{Alternate Experimental Settings}
In this section, we study other strategies that we could have used to quantify memorization.

\label{section:other_approaches}
\minisection{Random dataset sampling.}
In Section \ref{sec:alt_settings}, we explored what would happen if we instead chose a
truly random subset of the training data, where each sequence is sampled uniformly.
Specifically, we randomly sample $100{,}000$ sequences
from The Pile dataset of length $100$, $200$, $500$, and $1{,}000$;
prompt the model with the first $N-50$ tokens; and then
test for memorization by verifying if the model can emit the remaining $50$ tokens perfectly.    
In our analysis in 
Figures~\ref{fig:other_approaches_randomsize} and \ref{fig:other_approaches_randomlength}, we vary the size of the trained model and the
context length we provide it to understand how these factors impact memorization---but this time through prompting the models
with randomly sampled training sequences.
As expected, the absolute probability of memorization is much lower than in Figure~1 where we prompted models with training data from the sampled duplication-normalized subset.

We observe similar trends with model scale and context length as in our other results.
Larger models memorize more training examples than smaller models---and much more than the baseline GPT-2 model
that was not trained on The Pile.
Similarly, providing more context to a model increases
the likelihood we can discover memorization.
In Figure~\ref{fig:other_approaches_randomlength}, we prompt models with: $\text{prompt length} = \text{sequence length} - 50$. We see that the longer prompts are easier to predict correctly than shorter prompts. 
The baseline GPT-2 model is nearly twice as accurate on
sequences of length $1{,}000$ (prompt length $ = 950$) compared to sequences of length $100$ (prompt length $ = 50$).

\begin{figure*}
    \centering
    \includegraphics[width=\linewidth]{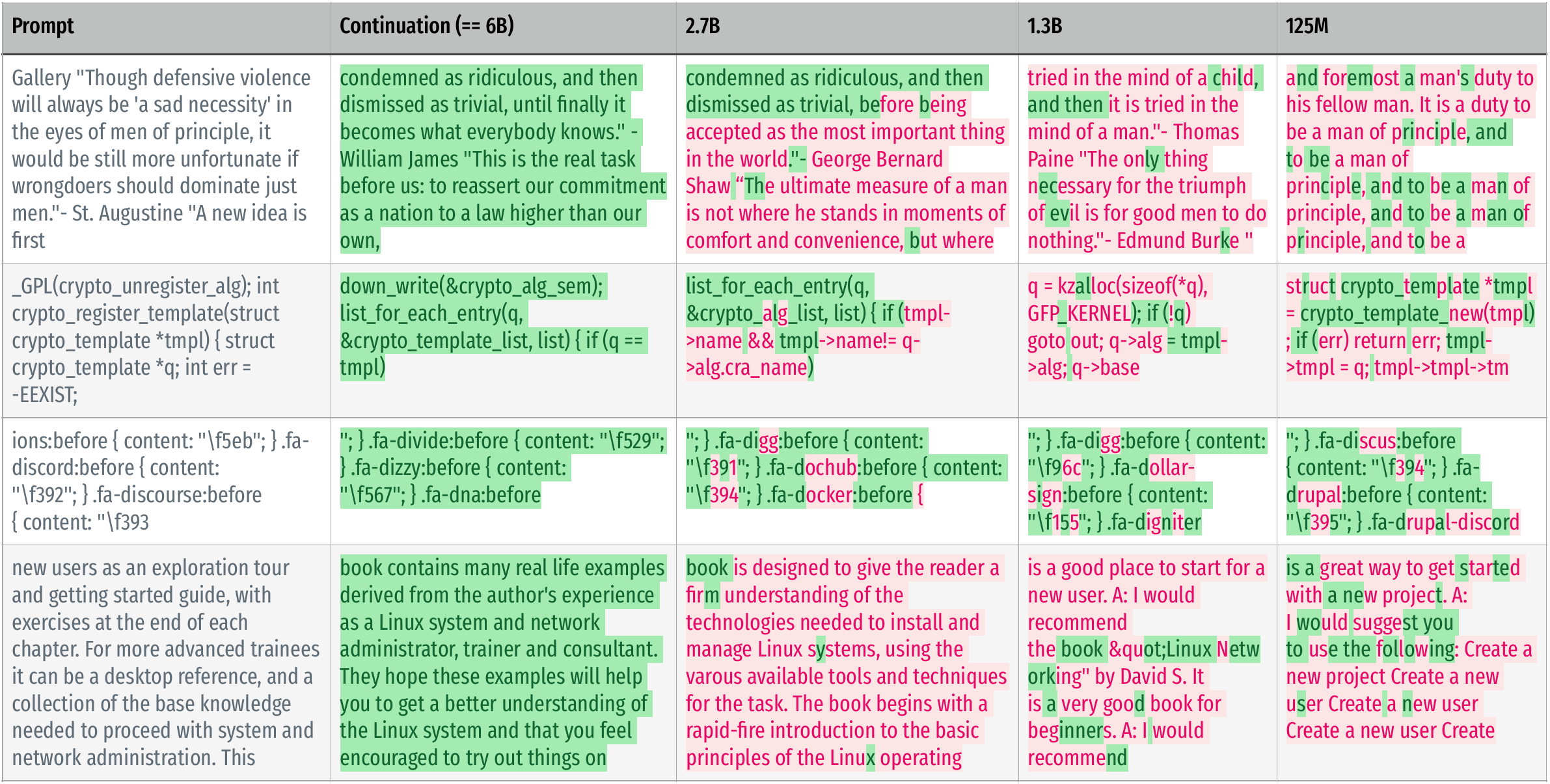}
    \caption{Text examples that are memorized by the 6B model, but not by smaller models. Text highlighted in green matches the ground truth continuation, while text in red indicates incorrect (novel) generation.}
    \label{fig:egs-mem-by-6b-not-smaller}

\end{figure*}

\minisection{Alternate definition of extractability.}
Our main experiments report a sequence as ``extractable'' if the model's generated continuation is identical to the true suffix within that training example.
This method is a loose lower bound on memorization.
Consider two sequences $x_1$, $x_2$ both contained in the training dataset.
Suppose these two sequences share the same prefix, and differ only in the final suffix;
that is, $x_1 = [p || s_1]$ and $x_2 = [p || s_2]$.
When we select $x_1$ and prompt the model on the prefix $p$, we will 
report ``success'' \emph{only if the output equals $s_1$}, but not if the output is $s_2$,
even though this is \emph{also} a form of memorization.

We now consider how our results would change if we instead checked that the generation $[p || f(p)]$ from a prompt $p$ was contained \emph{anywhere} in the training dataset. This gives a strictly larger measurement of memorization.
By comparing these two methods (checking for memorization within the ground truth continuation, and within the entire dataset), we can understand how the choice of measurement affects the results in our experiments.


Searching within the entire dataset finds more memorized content than comparing with the ground truth (Figure \ref{fig:other_approaches_search}). 
For examples at 100 repetitions $32.6\%$ of outputs are contained somewhere in the dataset but just $15.8\%$ match the ground truth continuation.
This difference becomes more pronounced as the number of repetitions increases. 
The maximum difference between these approaches is 28.4\%, at 2{,}200 repetitions.

We refrain from using this approach for our main experiments,
%
%
because this definition requires vastly larger computation resources; it requires
querying whether hundreds of thousands of sequences are contained in an 800GB
training dataset.
Therefore, to promote reproducability, the remainder of this paper continues with testing the generated suffix against the single expected training suffix.
%

\section{Text Memorized by Only Some Models}
\begin{table}[h]
    \small
    \centering\caption{The number of sequences memorized by one model, and not memorized by another.
    }
    \begin{tabular}{@{} l r r r r r @{}}
        \toprule
            &&  \multicolumn{4}{c}{Not Memorized By}\\
        \cmidrule{3-6}
         Model & Memorized & 125M & 1.3B & 2.7B & 6B\\
         \midrule
         125M & 4{,}812 &
         - & \cellcolor{blue!2}328 & \cellcolor{blue!1}295 & \cellcolor{blue!1}293\\
         
         1.3B & 10{,}391 &
         \cellcolor{blue!29}5{,}907 & - & \cellcolor{blue!6}1{,}205 & \cellcolor{blue!5}1{,}001\\
         
         2.7B & 12{,}148 &
         \cellcolor{blue!37}7{,}631 & \cellcolor{blue!14}2{,}962 & - & \cellcolor{blue!7}1{,}426\\
         
         6B & 14{,}792 &
         \cellcolor{blue!50}10{,}273 & \cellcolor{blue!26}5{,}402 & \cellcolor{blue!20}4{,}070 & - \\
         \bottomrule
    \end{tabular}
    \label{tab:memorized_by_model}
\end{table}

Table~\ref{tab:memorized_by_model} shows the total number of sequences that are memorized by
one model but not another.
Larger models have more uniquely memorized sequences, although every model has some memorization not shared by any other model.
(Even the $125$M model memorizes a few sequences that the $6$B model does not.)

\section{Memorization in OPT Models}

\begin{figure}[h]
    \centering
    \begin{subfigure}[a]{0.4\textwidth}
        \includegraphics[width=\textwidth]{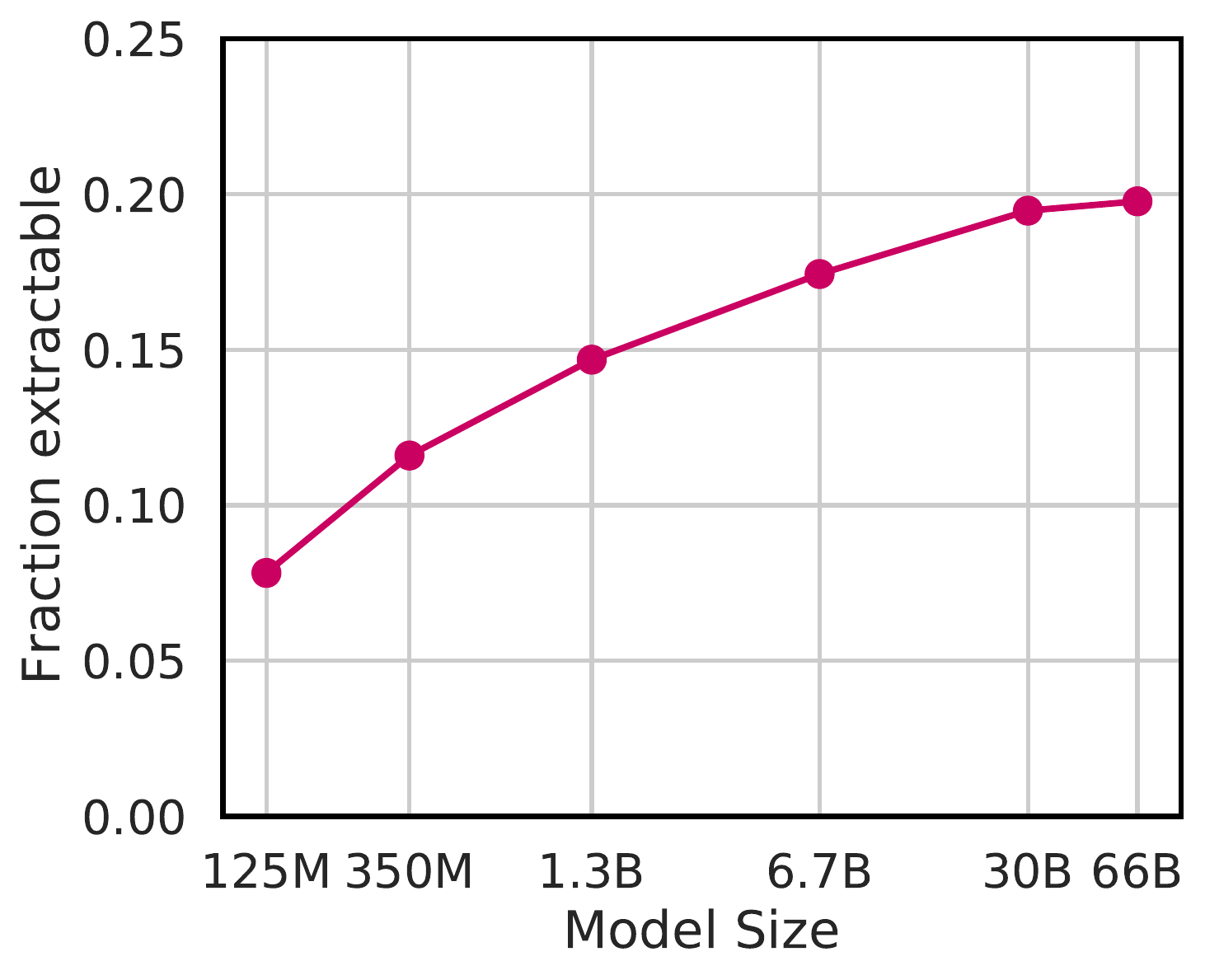}
    
        \caption{}
    \end{subfigure}
    \hfill
    \begin{subfigure}[a]{0.4\textwidth}
        \includegraphics[width=\textwidth]{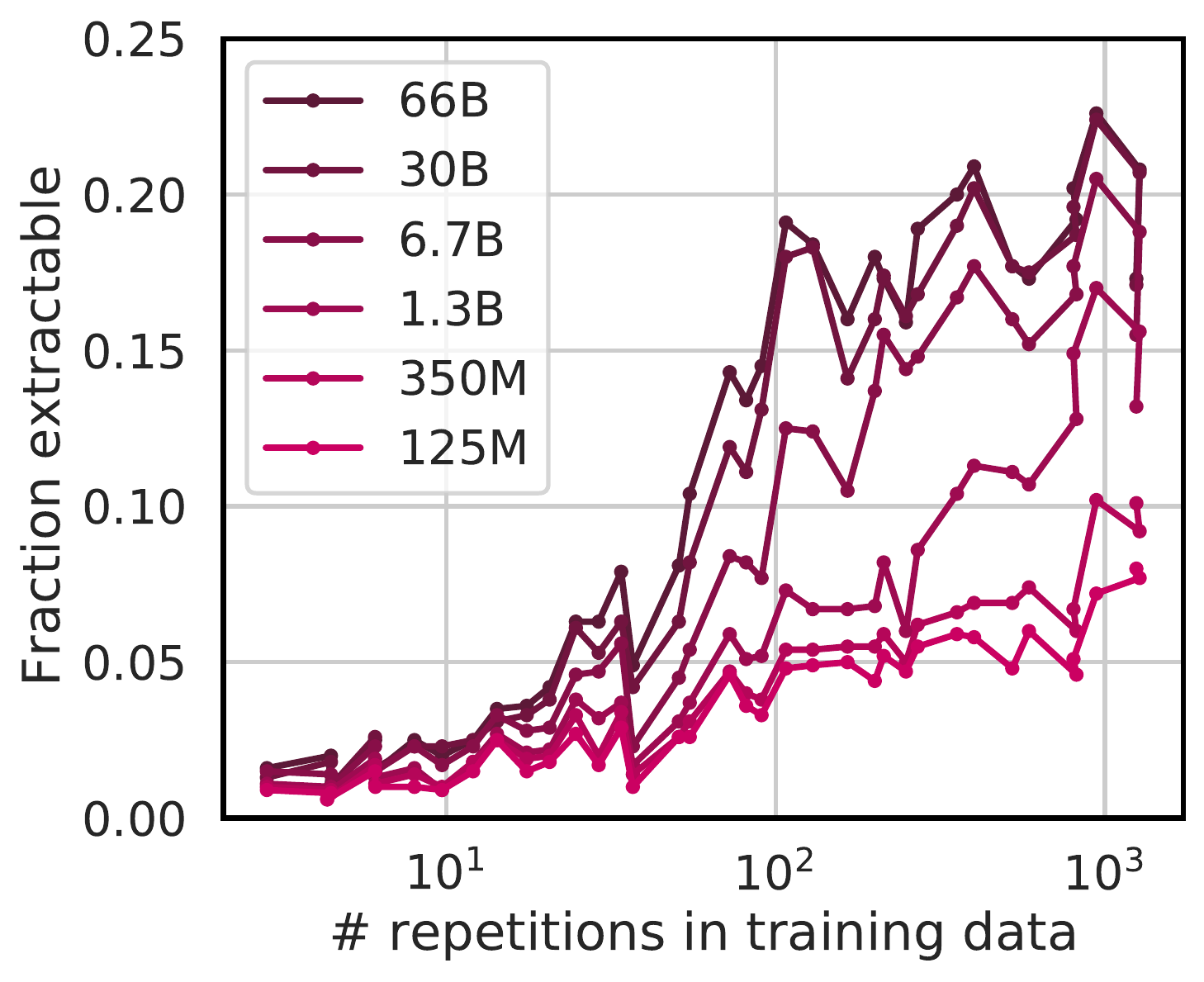} 
        \caption{}
    \end{subfigure}
    
\caption{We prompt OPT models with data sampled from their training set. We use a prompt length of 100 here. \textbf{(a)} Fraction of sequences extracted as a function of model scale. \textbf{(b)} Fraction of sequences extracted as the number of repetitions of that sequence in the training set increases.}
\label{fig:opt}
\end{figure}

\section{Examples of Memorized Texts}

\label{app:qualitative-examples}

\begin{figure}
    \centering
    \includegraphics[width=.8\linewidth]{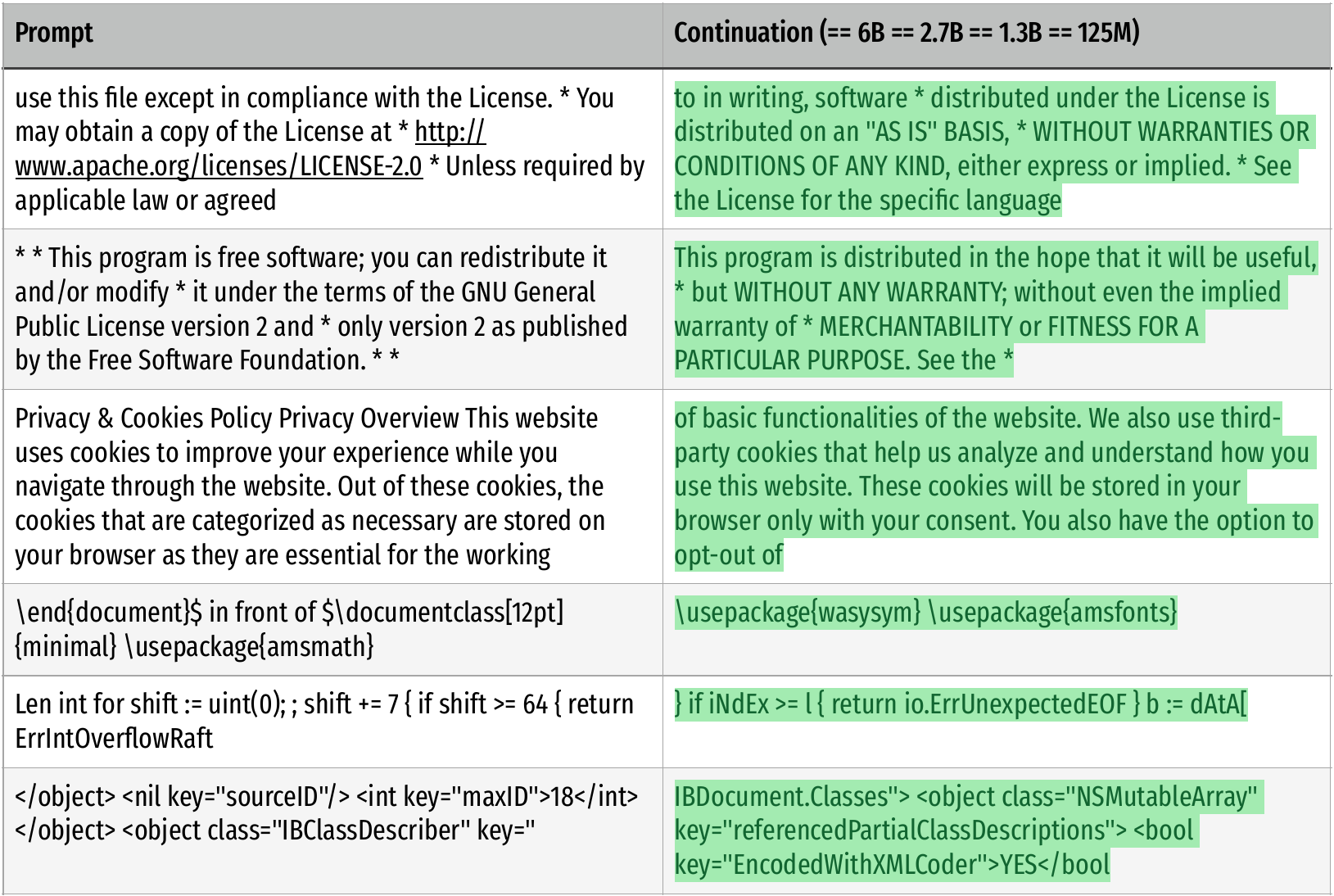}
    \caption{Text examples that are memorized by all the models: given 50-token prompts on the left, the next 50 tokens generated by all the models match the groundtruth continuation.}
    \label{fig:egs-mem-by-all}
\end{figure}

We show examples of texts that are memorized by different models. We consider the case of 50-token prompts and 50-token generation. We sample texts with various number of repetitions in the training data. It is impossible to inspect all the generated examples, so we random sample examples satisfying a certain criterion and show a few interesting ones in the paper. Figure~\ref{fig:egs-mem-by-all} lists examples that are memorized by models of \emph{all} sizes, in the sense that the 50-token generations match the groundtruth continuations of the prompts.

Figure~\ref{fig:egs-mem-by-6B} lists examples that are memorized by the 6B model but not by smaller ones. Specifically, the 50-token generations of the 6B model match the groundtruth continuations exactly, but the generations from the smaller models match \emph{neither} the groundtruth continuations of the prompted examples \emph{nor} any other training examples with the same prompts. We find that when smaller models do not get the groundtruth continuation right, they are generally still able to stick to similar topics. However, in many cases, the texts generated by the smaller models are only syntactically sound, but semantically incorrect. Figure~\ref{fig:egs-mem-by-6B-app1} and Figure~\ref{fig:egs-mem-by-6B-app2} show more examples.

In Figure~\ref{fig:egs-mem-by-125M} we show examples that are only memorized by the smallest model, using similar criterion as when we filter examples that are only memorized by the largest model. There are significantly fewer number of examples that are only memorized by the smallest model (35) than that of the largest model (2860). One of those examples (the first row of Figure~\ref{fig:egs-mem-by-125M}) is particularly interesting: the groundtruth continuation contains a typo due to formatting cutoff. While the smallest model memorized the typo, larger models try to fix the typo.

In Figure~\ref{fig:egs-mem-but-not-rep} and Figure~\ref{fig:egs-rep-but-not-mem} we show examples that are memorized but not heavily duplicated in the training set, and examples that are heavily duplicated but not memorized, respectively. Finally, we show examples that are memorized by GPT2-XL in Figure~\ref{fig:egs-mem-by-gpt2}.

\begin{figure*}
    \centering
    \includegraphics[width=.9\linewidth]{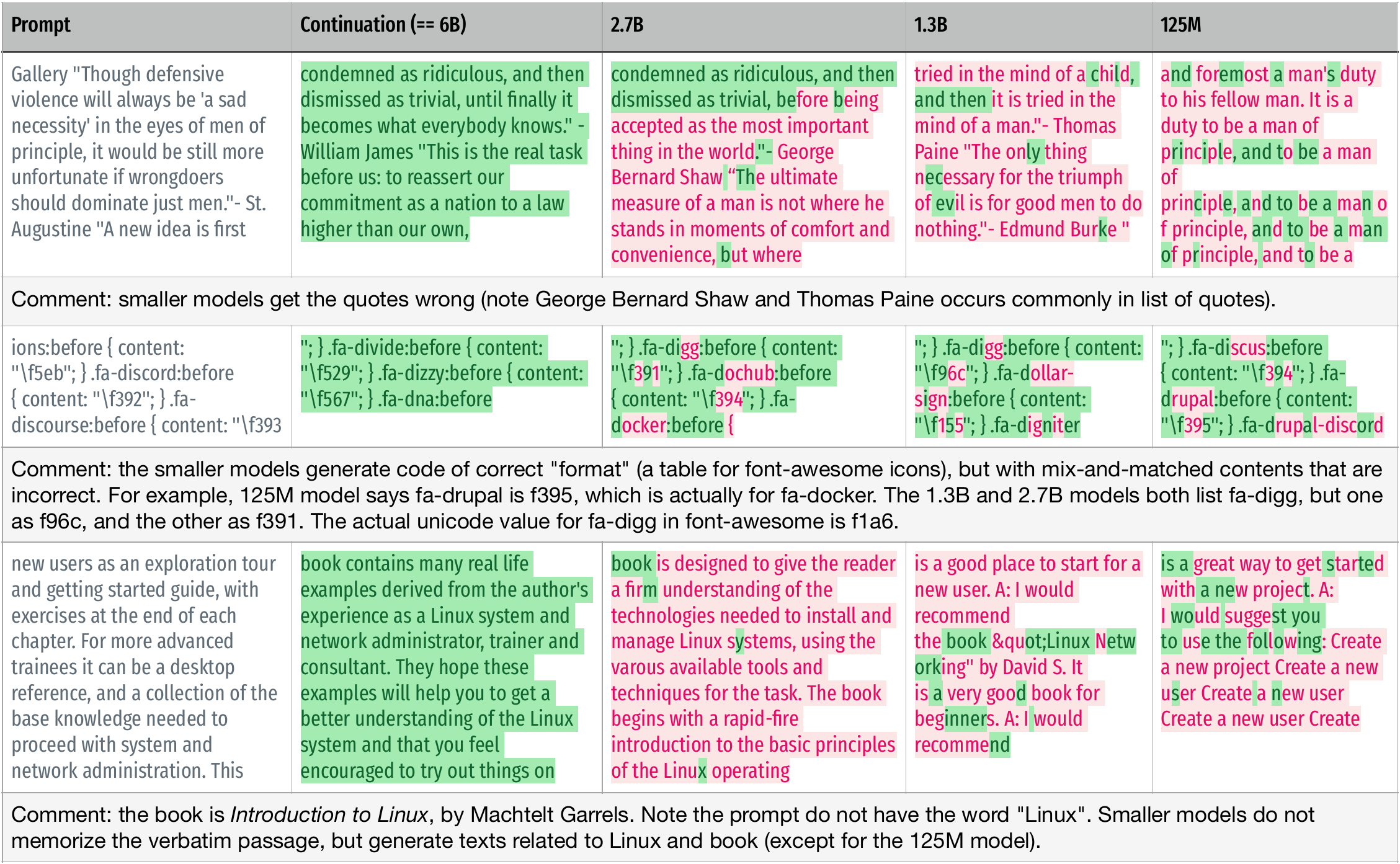}
    \caption{Text examples that are memorized by the 6B model (according to true-continuation match), but not memorized by smaller models (the generated texts do not match the true continuation, nor any other training examples). The first column shows the prompt. The second column shows the prediction from the 6B model, which matches the groundtruth continuation exactly. The remaining columns shows predictions from smaller models.}
    \label{fig:egs-mem-by-6B}
\end{figure*}

\begin{figure*}
    \centering
    \includegraphics[width=.9\linewidth]{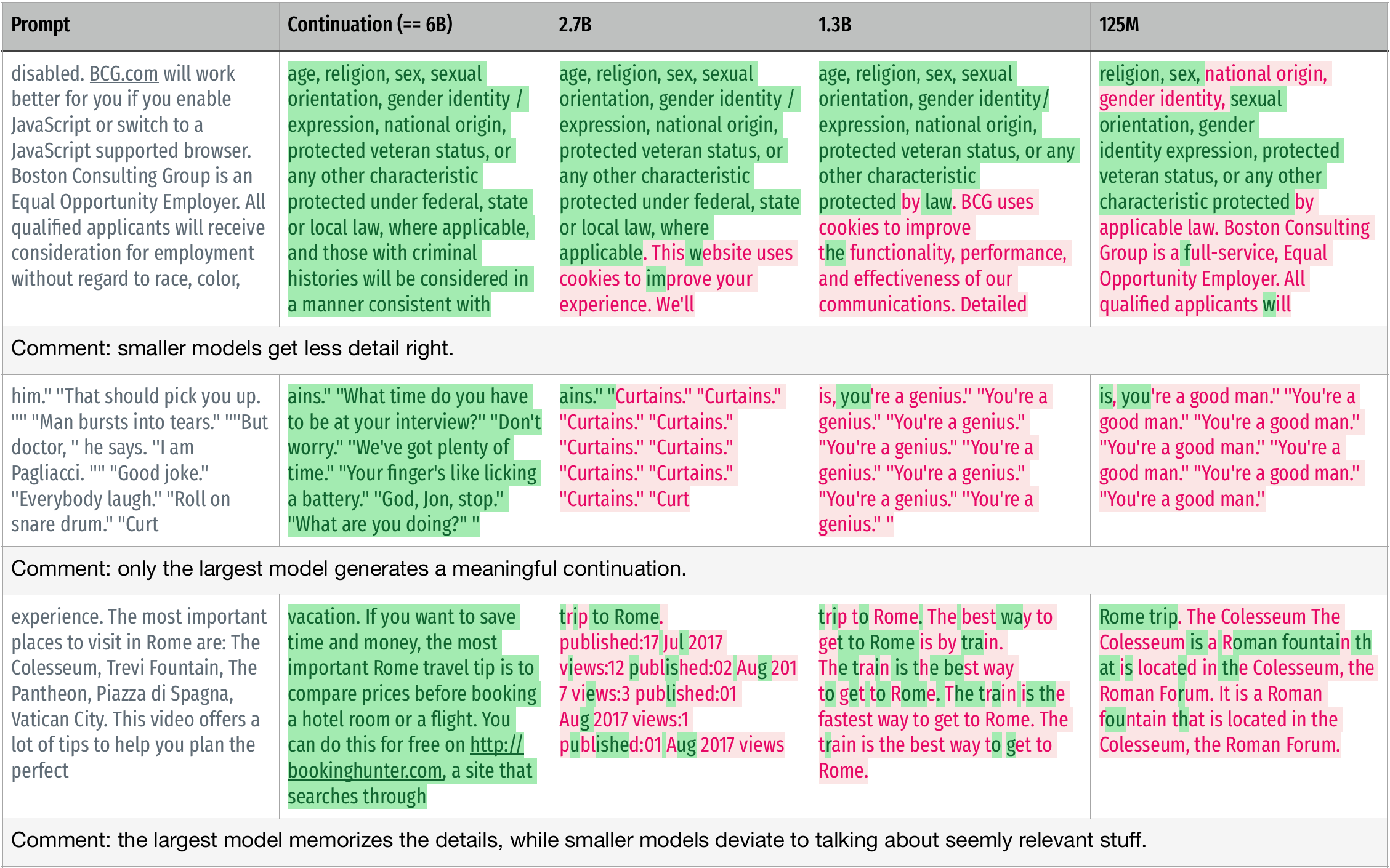}
    \caption{Continuation of Figure~\ref{fig:egs-mem-by-6B}.}
    \label{fig:egs-mem-by-6B-app1}
\end{figure*}

\begin{figure*}
    \centering
    \includegraphics[width=.9\linewidth]{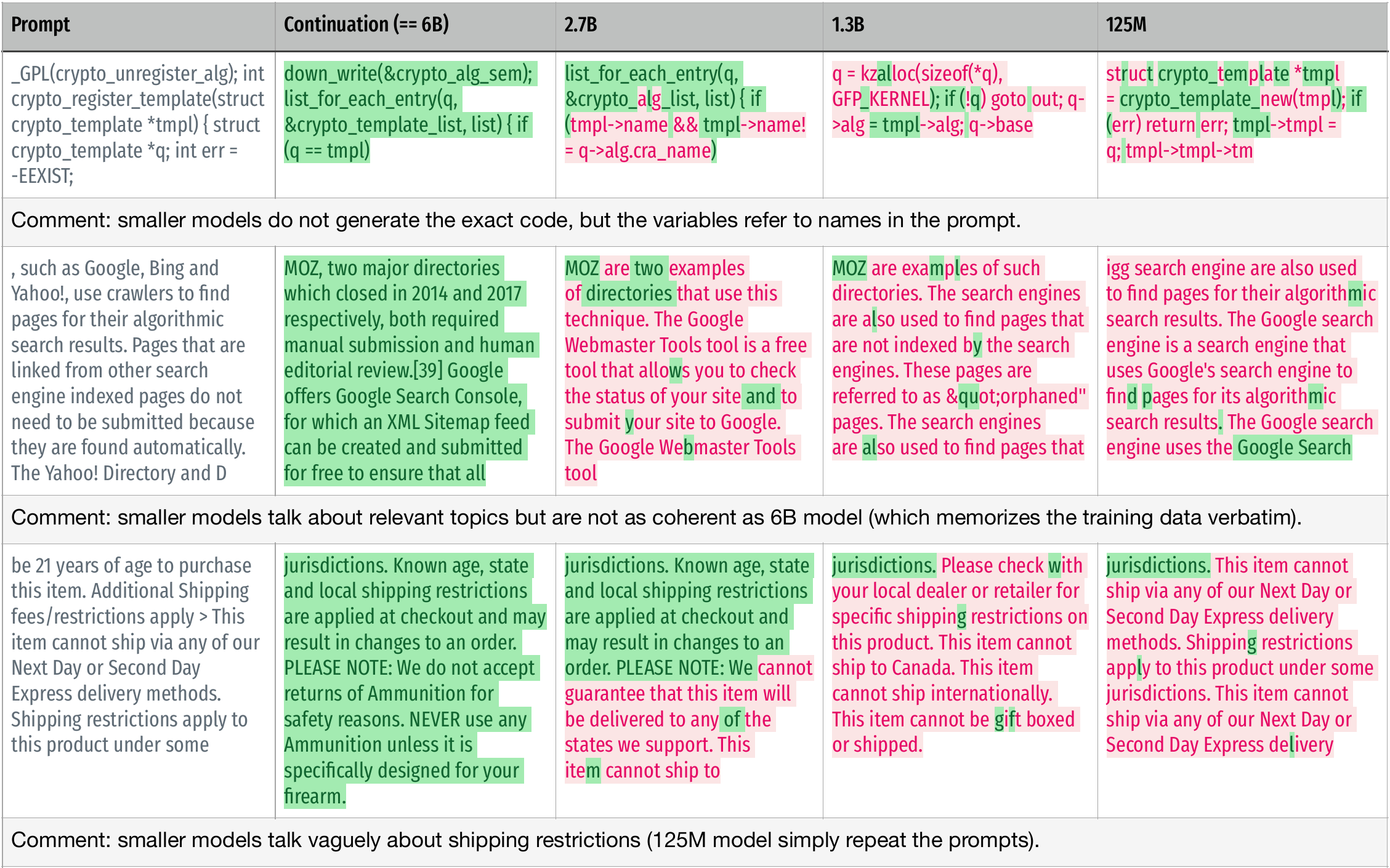}
    \caption{Continuation of Figure~\ref{fig:egs-mem-by-6B}.}
    \label{fig:egs-mem-by-6B-app2}
\end{figure*}

\begin{figure*}
    \centering
    \includegraphics[width=.9\linewidth]{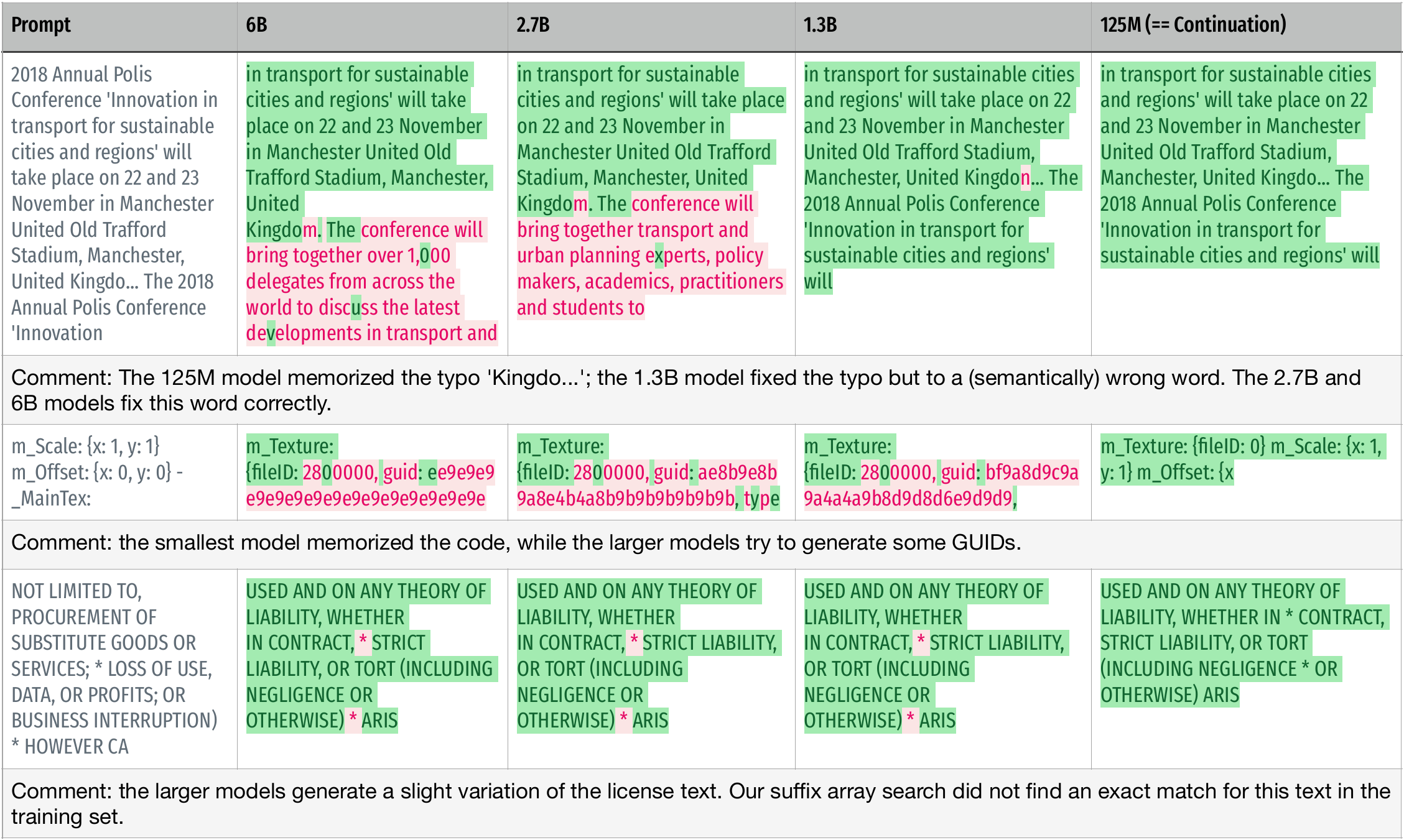}
    \caption{Text examples that are memorized by the 125M model (according to true-continuation match), but not memorized by larger models (the generated texts do not match the true continuation, nor any other training examples). The first column shows the prompt. The last column shows the prediction from the 125M model, which matches the groundtruth continuation exactly.}
    \label{fig:egs-mem-by-125M}
\end{figure*}

\begin{figure*}
    \centering
    \includegraphics[width=\linewidth]{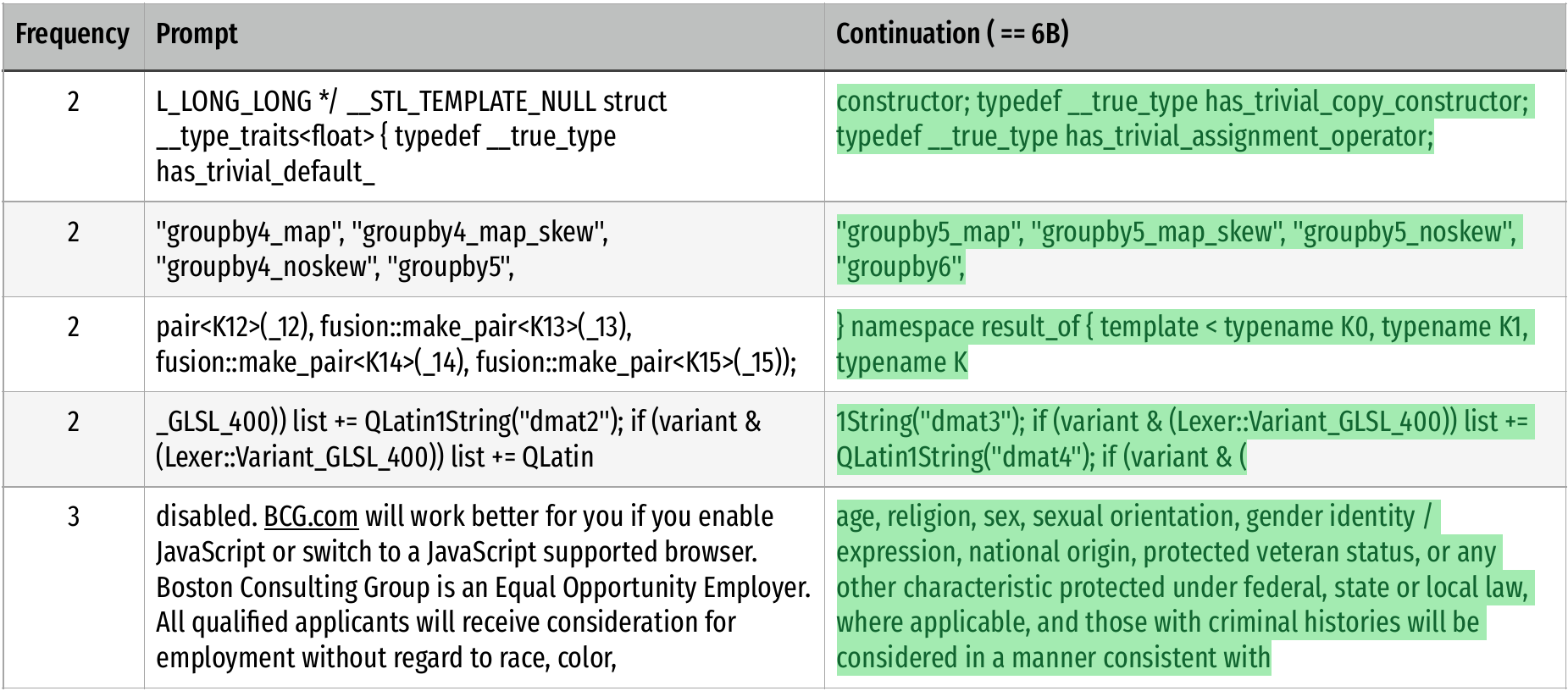}
    \caption{Text examples that are memorized but are not heavily duplicated in the training set. Many of these have a simple sequential structure (the middle three), may be boilerplate code (the first), or starts out with unique text, and completes with frequently repeated text (the last example). Overall, these are easily completed sequences.}
    \label{fig:egs-mem-but-not-rep}
\end{figure*}

\begin{figure*}
    \centering
    \includegraphics[width=\linewidth]{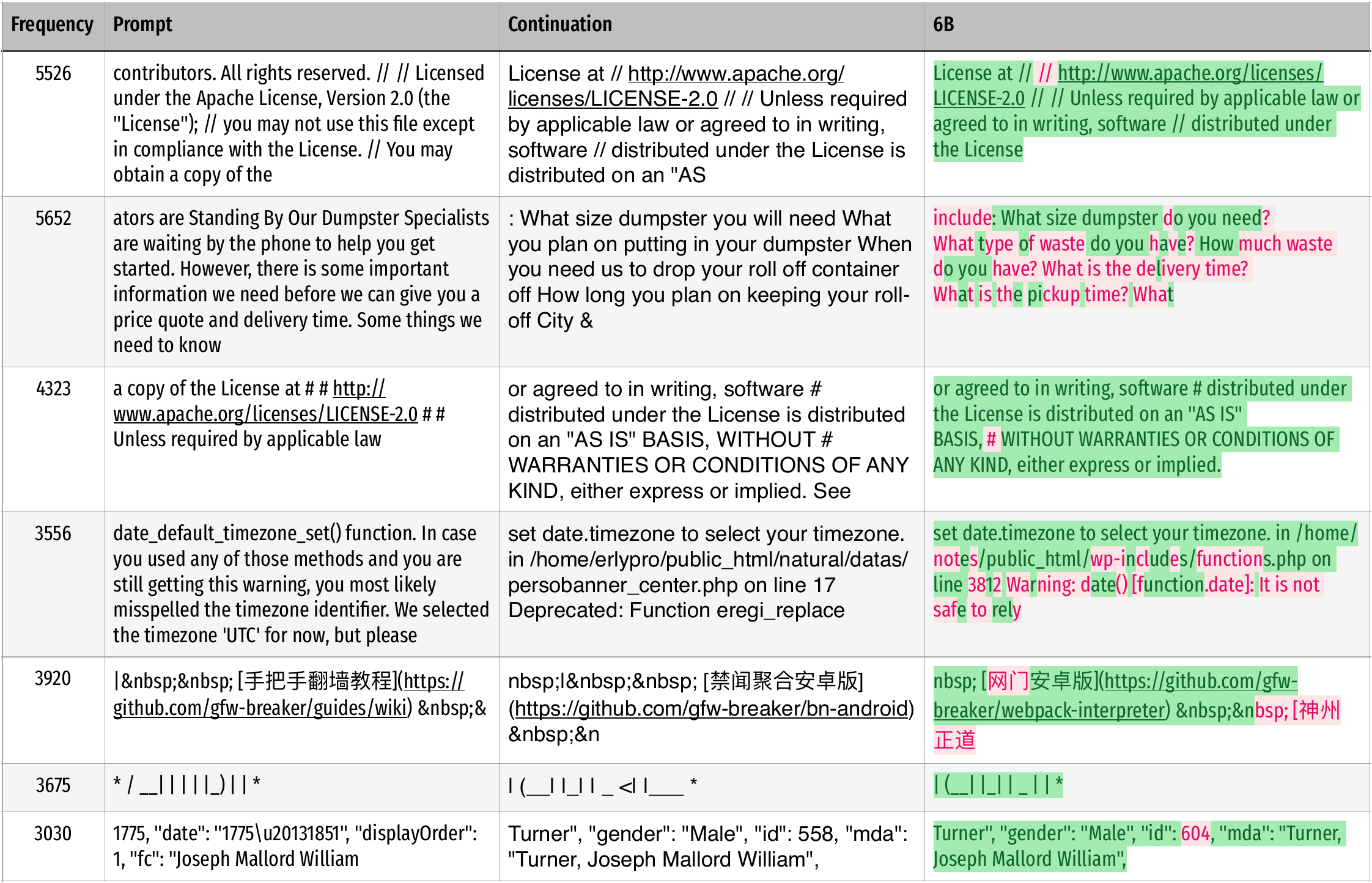}
    \caption{Text examples that are heavily replicated in the training set but not memorized. We find many examples which have slight differences with no semantic (English) meaning. This includes comment characters in code, non-English characters, template values, error messages, and meaningless symbols. We also surprisingly find a large number of slightly different but heavily repeated documents about dumpsters.}
    \label{fig:egs-rep-but-not-mem}
\end{figure*}

\begin{figure*}
    \centering
    \includegraphics[width=\linewidth]{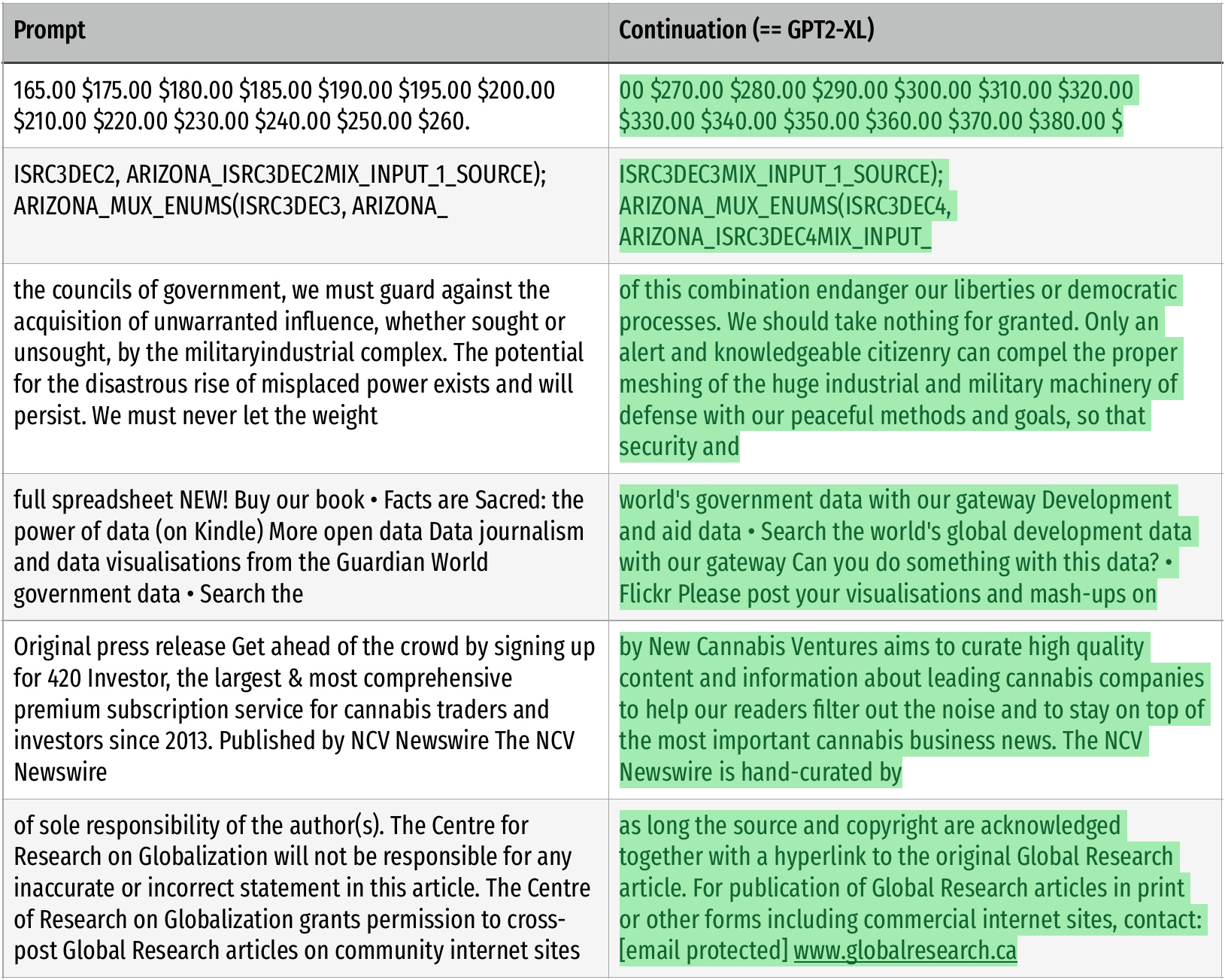}
    \caption{Text examples that are from The Pile and memorized by GPT2-XL. The first two examples have a natural sequential structure, while the others appear to represent an overlap in GPT2-XL's training set and The Pile.}
    \label{fig:egs-mem-by-gpt2}
\end{figure*}

\end{document}